\documentclass{egpubl}

 \JournalSubmission    
 \electronicVersion 

\ifpdf \usepackage[pdftex]{graphicx} \pdfcompresslevel=9
\else \usepackage[dvips]{graphicx} \fi

\PrintedOrElectronic

\usepackage{t1enc,dfadobe}

\usepackage{bm,overpic}
\usepackage{egweblnk}
\usepackage{cite}
\usepackage{hyperref}
\usepackage{amsmath}
\usepackage{amssymb}
\usepackage[export]{adjustbox}
\usepackage{wrapfig}
\usepackage[]{algorithm2e}
\usepackage{pgfplots}

\usepackage{pgf,tikz}
\usepackage{tkz-euclide}
\usepackage{tkz-graph}

\title[Partial Functional Correspondence]%
      {Partial Functional Correspondence} 

\newcommand{\rev}[1]{#1}
\newcommand{\eg}{\emph{e.g.}}
\newcommand{\ie}{\emph{i.e.}}
\newcommand{\etal}{\emph{et al.}}

\newcommand{\vct}[1]{\ensuremath{\mathbf{#1}}}
\newcommand{\T}{\ensuremath{^\top}}
\newcommand{\diag}{\operatorname{diag}}

\newcommand{\C}{\mathbf{C}}
\newcommand{\A}{\mathbf{A}}
\newcommand{\B}{\mathbf{B}}
\newcommand{\D}{\mathbf{H}}
\newcommand{\G}{\mathbf{G}}
\newcommand{\F}{\mathbf{F}}
\newcommand{\M}{\mathcal{M}}
\newcommand{\N}{\mathcal{N}}

\newlength\figureheight 
\newlength\figurewidth

\author[E. Rodol{\`a} et al.]
       {E. Rodol{\`a}$^1$, L. Cosmo$^2$, M. M. Bronstein$^3$, A. Torsello$^2$, D. Cremers$^1$
        \\
        $^1$TU Munich, Germany \,\,\,\,\, $^2$University of Venice, Italy \,\,\,\,\, $^3$University of Lugano, Switzerland
       }

\begin{document}

\teaser{
 \includegraphics[width=\linewidth]{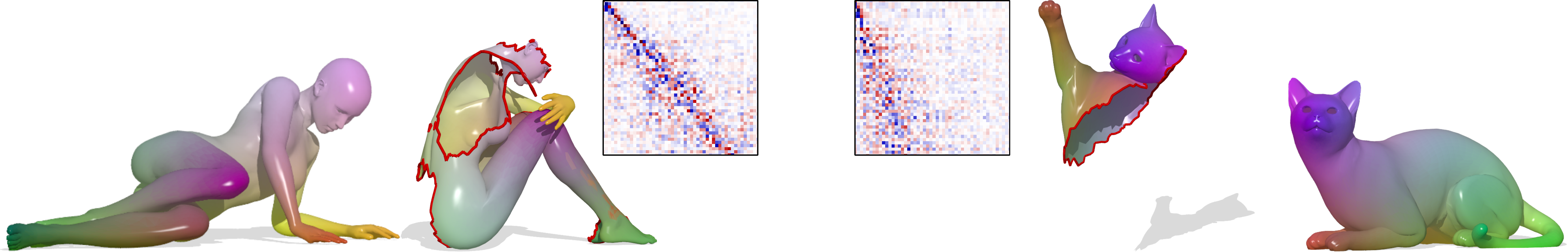}
 \centering
  \caption{Partial functional correspondence between two pairs of shapes with large missing parts. For each pair we show the matrix $\mathbf{C}$ representing the functional map in the spectral domain, and the action of the map by transferring colors from one shape to the other. The special slanted-diagonal structure of $\mathbf{C}$ induced by the partiality transformation is first estimated from spectral properties of the two shapes, and then exploited to drive the matching process.}
\label{fig:teaser}
}

\maketitle

\begin{abstract}
In this paper, we propose a method for computing  partial functional correspondence between non-rigid shapes. We use perturbation analysis to show how removal of shape parts changes the Laplace-Beltrami eigenfunctions, and exploit it as a prior on the spectral representation of the correspondence. Corresponding parts are optimization variables in our problem and are used to weight the functional correspondence; we are looking for the largest and most regular (in the Mumford-Shah sense) parts that minimize correspondence distortion. We show that our approach can cope with very challenging correspondence settings.
\begin{classification} 
\CCScat{Computer Graphics}{I.3.5}{Computational Geometry and Object Modeling}{Shape Analysis}
\end{classification}

\end{abstract}


\section{Introduction}

The problem of shape correspondence is one of the most fundamental problems in computer graphics and geometry processing, with a plethora of applications ranging from texture mapping to animation \cite{bronstein2006generalized,DBLP:journals/cgf/KimLCF10,DBLP:journals/tog/KimLF11,van2011survey}. 
A particularly challenging setting is that of {\em non-rigid correspondence}, where the shapes in question are allowed to undergo deformations, which are typically assumed to be approximately isometric (such a model appears to be good for, \eg, human body poses). Even more challenging is {\em partial correspondence}, where one is shown only a subset of the shape and has to match it to a deformed full version thereof. Partial correspondence problems arise in numerous applications that involve real data acquisition by 3D sensors, which inevitably lead to missing parts due to occlusions or partial view. 

\paragraph*{Related work.}


For rigid partial correspondence problems, arising \eg, in 3D scan completion applications, many versions of regularized iterative closest point (ICP) approaches exist, see for example \cite{aiger20084,albarelli-pr15}. 
Attempts to extend these ideas to the non-rigid case in the form of non-rigid or piece-wise rigid ICP have been explored in recent years \cite{li08}. By nature of the ICP algorithm, these methods rely on the assumption that the given shapes can be placed in approximate rigid alignment to initiate the matching process. As a result, they tend to work well under small deformations (\eg, when matching neighboring frames of a sequence), but performance deteriorates quickly when this assumption does not hold.

For the non-rigid setting, several metric approaches centered around the notion of minimum distortion correspondence \cite{bronstein2006generalized} have been proposed. Bronstein~\etal \cite{bronstein2008not,bronstein2009partial} combine metric distortion minimization with optimization over matching parts, showing an algorithm that simultaneously seeks for a  correspondence and maximizes the {\em regularity} of corresponding parts in the given shapes. 
Rodol\`{a}~\etal \cite{rodola12} subsequently relaxed the regularity requirement by allowing sparse correspondences, and later introduced a mechanism to explicitly control the degree of sparsity of the solution \cite{rodola13iccv}. Finally, in \cite{sahilliouglu2014partial} the authors proposed a voting-based formulation to match shape extremities, which are assumed to be preserved by the partiality transformation. Being based on spectral features and metric preservation, the accuracy of the aforementioned methods suffers at high levels of partiality, where the computation of these quantities becomes unreliable due to boundary effects and meshing artifacts. Furthermore, these methods suffer from high computational complexity and generally provide only a {\em sparse correspondence}. 

Pokrass~\etal \cite{pokrass13-nmtma} proposed a descriptor-based partial matching approach where the optimization over parts is done to maximize the matching of bags of local descriptors. The main drawback of this approach is that it only finds similar parts, without providing a correspondence between them. 
Windheuser~\etal \cite{wind11} formulated the shape matching problem as one of seeking minimal surfaces in the product space of two given shapes; the formulation notably allows for a linear programming discretization and provides guaranteed continuous and orientation-preserving solutions. The method was shown to work well with partial shapes, but requires watertight surfaces as input (\eg, via hole filling). 
Brunton~\etal \cite{Brunton201470} used alignment of tangent spaces for partial correspondence. In their method, a sparse set of correspondences is first computed by matching feature descriptors; the matches are then propagated in an isometric fashion so as to cover the largest possible regions on the two shapes. Since the quality of the final solution directly depends on the initial matches, the method is understood as a ``densification'' method to complement other sparse approaches.
Other recent works include the design of robust descriptors for partial matching \cite{kaick13}. In the context of collections of shapes, partial correspondence has been considered in \cite{van2011prior,chen14icml,cosmo15}.

All the aforementioned works are based on the notion of point-wise correspondence between shapes. 
Recently, Ovsjanikov~\etal \cite{ovsjanikov12} proposed the {\em functional maps} framework, in which shape correspondence is modeled as a linear operator between spaces of functions on the shapes. 
The main advantage of functional maps is that finding correspondence boils down to a simple algebraic problem, as opposed to difficult combinatorial-type problems arising in, \eg, the computation of minimum-distortion maps. 
While several recent works showed that functional maps can be made resilient to missing parts or incomplete data \shortcite{DBLP:journals/tog/HuangWG14,kovnatsky15}, overall this framework is not suitable for dealing with partial correspondence. 

%


\paragraph*{Contribution.} 
In this paper, we propose an extension to the functional correspondence framework to allow dealing with partial correspondence. 
Specifically, we consider a scenario of matching a part of a deformed shape to some full model. Such scenarios are very common for instance in robotics applications, where one has to match an object acquired by means of a 3D scanner (and thus partially occluded) with a reference object known in advance. 
We use an explicit part model over which optimization is performed as in \cite{bronstein2008not,bronstein2009partial}, as well as a regularization on the spectral representation of the functional correspondence accounting for a special structure of the Laplacian eigenfunctions as a result of part removal. Theoretical study of this behavior based on perturbation analysis of Laplacian matrices is another contribution of our work. 
We show experimentally that the proposed approach allows dealing with very challenging partial correspondence settings; further, we introduce a new benchmark to evaluate deformable partial correspondence methods, consisting of hundreds of shapes and ground-truth information.


The rest of the paper is organized as follows. 
In Section \ref{sec:bg}, we review the basic concepts in the spectral geometry and describe the functional correspondence approach. 
Section \ref{sec:perturb} studies the behavior of Laplacian eigenfunctions in the case of missing parts, motivating the regularizations used in the subsequent sections. 
Section \ref{sec:method} introduces our partial correspondence model, and Section \ref{sec:impl} describes its implementation details. 
Section \ref{sec:exp} presents experimental results, and finally, Section \ref{sec:conclusion} concludes the paper.

\section{Background}
\label{sec:bg}

In this paper, we model shapes as compact connected \rev{2-manifolds} $\mathcal{M}$, possibly with boundary $\partial\mathcal{M}$. 
%
Given $f, g : {\mathcal{M} \rightarrow \mathbb{R}}$ some real scalar fields on the manifold, we define the standard inner product $\langle f, g\rangle_{\mathcal{M}} = \int_{\mathcal{M}}f(x)g(x) dx$, where integration is done using the area element induced by the Riemannian metric.  
 We denote by $L^2(\mathcal{M}) = \{ {f: \mathcal{M} \rightarrow\mathbb{R}} ~|~  \langle f, f\rangle_{\mathcal{M}} <\infty \}$ the space of square-integrable functions on $\mathcal{M}$.

The {\em intrinsic gradient} $\nabla_{\mathcal{M}}f$ and the positive semi-definite {\em Laplace-Beltrami operator} $\Delta_{\mathcal{M}}f = -\mathrm{div}_{\mathcal{M}}( \nabla_{\mathcal{M}}f )$ generalize the notions of gradient and Laplacian to manifolds. 
  %
 %
 The Laplace-Beltrami operator admits an eigen-decomposition 
 \begin{eqnarray}
\Delta_{\mathcal{M}} \phi_i(x) = \lambda_i \phi_i(x)  & \,\,\,\,\,& x \in \mathrm{int}(\mathcal{M}) \\
\langle \nabla_{\mathcal{M}} \phi_i(x) , \hat{n}(x) \rangle = 0 &\,\,\,\,\,& x \in \partial\mathcal{M}, 
\label{eq:neumann}
\end{eqnarray}
with homogeneous Neumann boundary conditions~(\ref{eq:neumann}) if $\mathcal{M}$ has a boundary  (here $\hat{n}$ denotes the normal vector to the boundary), where $0 = \lambda_1 <\lambda_2 \leq \hdots$ are eigenvalues and  $\phi_1, \phi_2, \hdots$ are the corresponding eigenfunctions (or eigenvectors).  
The eigenfunctions form an orthonormal basis on $L^2(\mathcal{M})$, \ie, $\langle \phi_i, \phi_j\rangle_{\mathcal{M}} = \delta_{ij}$, generalizing the classical Fourier analysis: a function $f\in L^2(\mathcal{M})$ can be expanded into the {\em Fourier series} as 
\begin{eqnarray}
\label{eq:fourier}
f(x) &=& \sum_{i\geq 1}  \langle f, \phi_i\rangle_{\mathcal{M}} \phi_i(x)\,.
\end{eqnarray}

\paragraph*{Functional correspondence.}
Let us be now given two manifolds, $\mathcal{N}$ and $\mathcal{M}$. 
Ovsjanikov \etal \shortcite{ovsjanikov12} proposed modeling {\em functional correspondence} between shapes as a linear operator $T: L^2(\mathcal{N}) \rightarrow L^2(\mathcal{M})$. 
One can easily see that classical vertex-wise correspondence is a particular setting where $T$ maps delta-functions to delta-functions. 

Assuming to be given two orthonormal bases $\{\phi_i\}_{i\geq 1}$ and $\{\psi_i\}_{i\geq 1}$ on $L^2(\mathcal{N})$ and $L^2(\mathcal{M})$ respectively, the functional correspondence can be expressed w.r.t. to these bases as follows: 
\begin{eqnarray}
\label{eq:funcorr1}
Tf &=& 
T \sum_{i\geq 1} \langle f, \phi_i \rangle_{\mathcal{N}} \phi_i 
= \sum_{i\geq 1} \langle f, \phi_i \rangle_{\mathcal{N}} T\phi_i \nonumber\\\label{eq:cc}
&=& \sum_{ij\geq 1} \langle f, \phi_i \rangle_{\mathcal{N}} 
\underbrace{\langle T\phi_i, \psi_j \rangle_{\mathcal{M}}}_{c_{ij}} \psi_j\,,
\end{eqnarray}
%
Thus, $T$ amounts to a linear transformation of the Fourier coefficients of $f$ from basis $\{\phi_i\}_{i\geq 1}$ to basis $\{\psi_i\}_{i\geq 1}$, which is captured by the coefficients $c_{ij}$. 
Truncating the Fourier series~(\ref{eq:funcorr1}) at the first $k$ coefficients, one obtains a rank-$k$ approximation of $T$, represented in the bases  $\{\phi_i, \psi_i\}_{i\geq 1}$ as a $k\times k$ matrix $\mathbf{C} = (c_{ij})$.  

In order to compute $\mathbf{C}$, Ovsjanikov \etal \shortcite{ovsjanikov12} assume to be given a set of $q$ corresponding functions $\{ f_1, \hdots, f_q \} \subseteq L^2(\mathcal{N})$ and $\{ g_1, \hdots, g_q\} \subseteq L^2(\mathcal{M})$. 
%
Denoting by $a_{ij} = \langle f_j, \phi_i \rangle_{\mathcal{N}}$ and $b_{ij} = \langle g_j, \psi_i \rangle_{\mathcal{M}}$ the $k\times q$ matrices of the respective Fourier coefficients, functional correspondence boils down to the linear system 
\begin{eqnarray}
\label{eq:cab}
\mathbf{C} \mathbf{A} &=& \mathbf{B}\,.
\end{eqnarray}
 If $q\geq k$, the system~(\ref{eq:cab}) is (over-)determined and is solved in the least squares sense to find $\mathbf{C}$.

\paragraph*{Structure of C.} 
We note that the coefficients $\mathbf{C}$ depend on the choice of the bases. In particular, it is convenient to use the eigenfunctions of the Laplace-Beltrami operators of $\mathcal{N}$ and $\mathcal{M}$ as the bases $\{\phi_i, \psi_i\}_{i\geq 1}$; truncating the series at the first $k$ coefficients has the effect of `low-pass' filtering thus producing smooth correspondences. In the following, this will be our tacit basis choice.  

Furthermore, note that the system~(\ref{eq:cab}) has $qk$ equations and $k^2$ variables. However, in many situations the actual number of variables is significantly smaller, as $\mathbf{C}$ manifests a certain structure which can be taken advantage of. 
In particular, if $\mathcal{N}$ and $\mathcal{M}$ are isometric and have simple spectrum (\ie, the Laplace-Beltrami eigenvalues have no multiplicity), then $T \phi_i = \pm \psi_i$, or in other words, $c_{ij} = \pm \delta_{ij}$. 
In more realistic scenarios (approximately isometric shapes), the matrix $\mathbf{C}$ would manifest a funnel-shaped structure, with the majority of elements distant from the diagonal close to zero. 

\paragraph*{Discretization.} 

\begin{figure}[bh!]
\centering
%
%
  \begin{overpic}
  [trim=0cm 0cm 0cm 0cm,clip,width=0.9\linewidth]{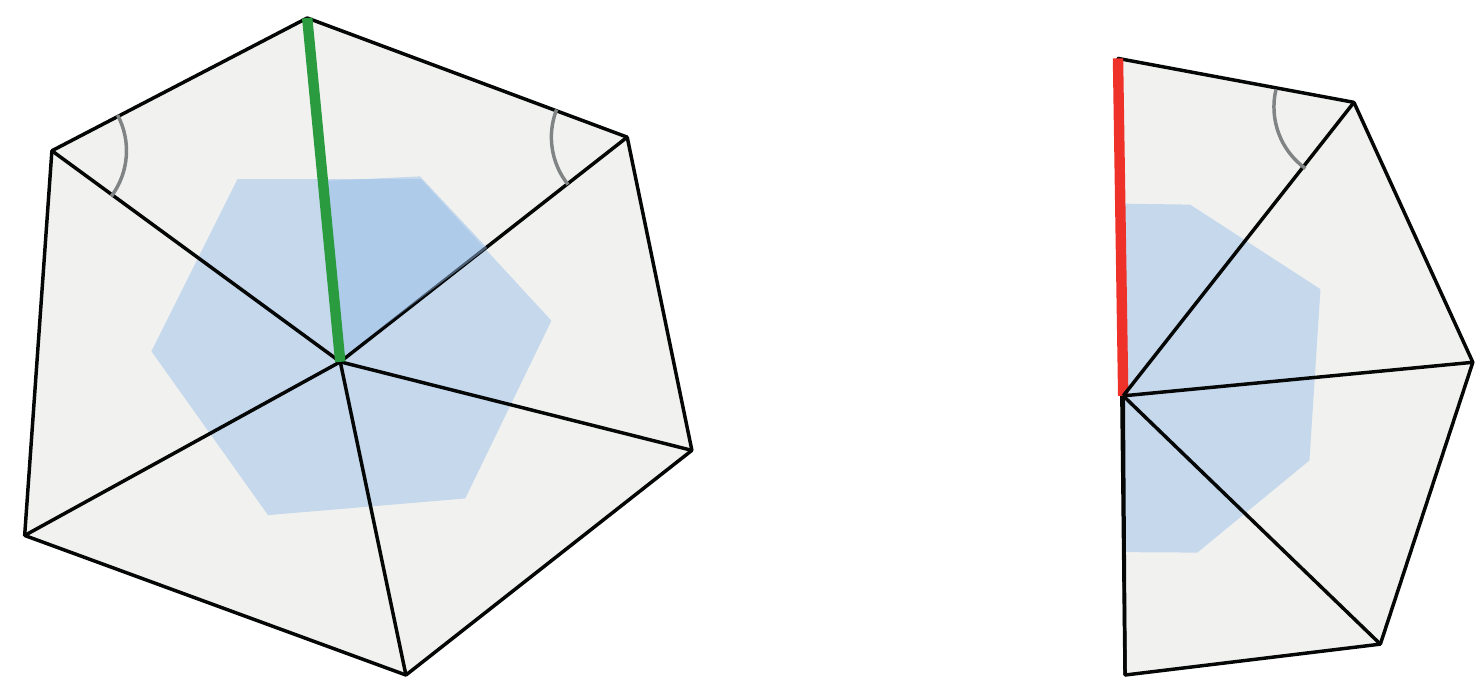}
  \put(21,19){\footnotesize $i$}
  \put(20,47){\footnotesize $j$}
  \put(43,37){\footnotesize $k$}
  \put(1,36){\footnotesize $h$}
  \put(16.5,35.5){\footnotesize $w_{ij}$}
  \put(23.5,30){\footnotesize $\tfrac{1}{3}s_{ijk}$}
  \put(32.5,36.5){\footnotesize $\alpha_{ij}$}
  \put(9,35.5){\footnotesize $\beta_{ij}$}
  \put(81.5,36.5){\footnotesize $\alpha_{ij}$}
  \put(73.5,19){\footnotesize $i$}
  \put(74,45){\footnotesize $j$}
  \put(92.5,39.5){\footnotesize $k$}
  \end{overpic}
\caption{Discretization of the Laplace-Beltrami operator on a triangular mesh for interior edges (green, left) and boundary edges (red, right). 
}
\label{fig:cot_weights}
\end{figure}

In the discrete setting, the manifold $\mathcal{N}$ is sampled at $n$ points $x_1, \hdots, x_n$ which are connected by edges $E = E_\mathrm{i} \cup E_\mathrm{b}$ and faces $F$, forming a manifold triangular mesh $(V,E,F)$.  We denote by $E_\mathrm{i}$ and $E_\mathrm{b}$ the interior and boundary edges respectively. 
%
%
A function on the manifold is represented by an $n$-dimensional vector $\mathbf{f} = (f(x_1), \hdots, f(x_n))^\top$. 
%
%
%
The discretization of the Laplacian takes the form of an $n\times n$ sparse matrix $\mathbf{L} = -\mathbf{S}^{-1}\mathbf{W}$ using the classical cotangent formula \cite{macneal1949solution,duffin1959distributed,pinkall1993computing},
\begin{eqnarray}
\label{eq:cotan}
w_{ij} & = & \left\{ 
		\begin{array}{lc}
			(\cot \alpha _{ij} + \cot \beta _{ij})/2 &   ij \in E_\mathrm{i};  \\		
			(\cot \alpha _{ij})/2 &   ij \in E_\mathrm{b};  \\		
			-\sum_{k\neq i} w_{ik}    & i = j; \\
			0 & \mathrm{else}; 
		\end{array}
\right.
\end{eqnarray}
where $\mathbf{S} = \mathrm{diag}(s_1, \hdots, s_n)$, $s_i = \frac{1}{3} \sum_{jk: ijk \in F} s_{ijk}$ denotes the local area element 
at vertex $i$, $s_{ijk}$ denotes the area of triangle $ijk$, and $\alpha_{ij}, \beta_{ij}$ denote the angles $\angle ikj, \angle jhi$ of the triangles sharing the edge $ij$ (see Fig.~\ref{fig:cot_weights}). 

%


The first $k$ eigenfunctions and eigenvalues of the Laplacian are computed by performing the generalized eigen-decomposition  $\mathbf{W} \bm{\Phi} = \mathbf{S}\bm{\Phi}\bm{\Lambda}$, where $\bm{\Phi} = (\bm{\phi}_1, \hdots, \bm{\phi}_k)$ is an $n\times k$ matrix containing as columns the discretized eigenfunctions and $\bm{\Lambda} = \mathrm{diag}(\lambda_1, \hdots, \lambda_k)$ is the diagonal matrix of the corresponding eigenvalues. 
The computation of Fourier coefficients is performed by $\mathbf{a} = \bm{\Phi}^\top \mathbf{S} \mathbf{f}$.

\newcommand{\vspacer}[1]{\rule{0pt}{#1}}

\begin{figure}[b]
  \centering
		\begin{overpic}
		[trim=0cm 0cm 0cm 0cm,clip,width=1\linewidth]{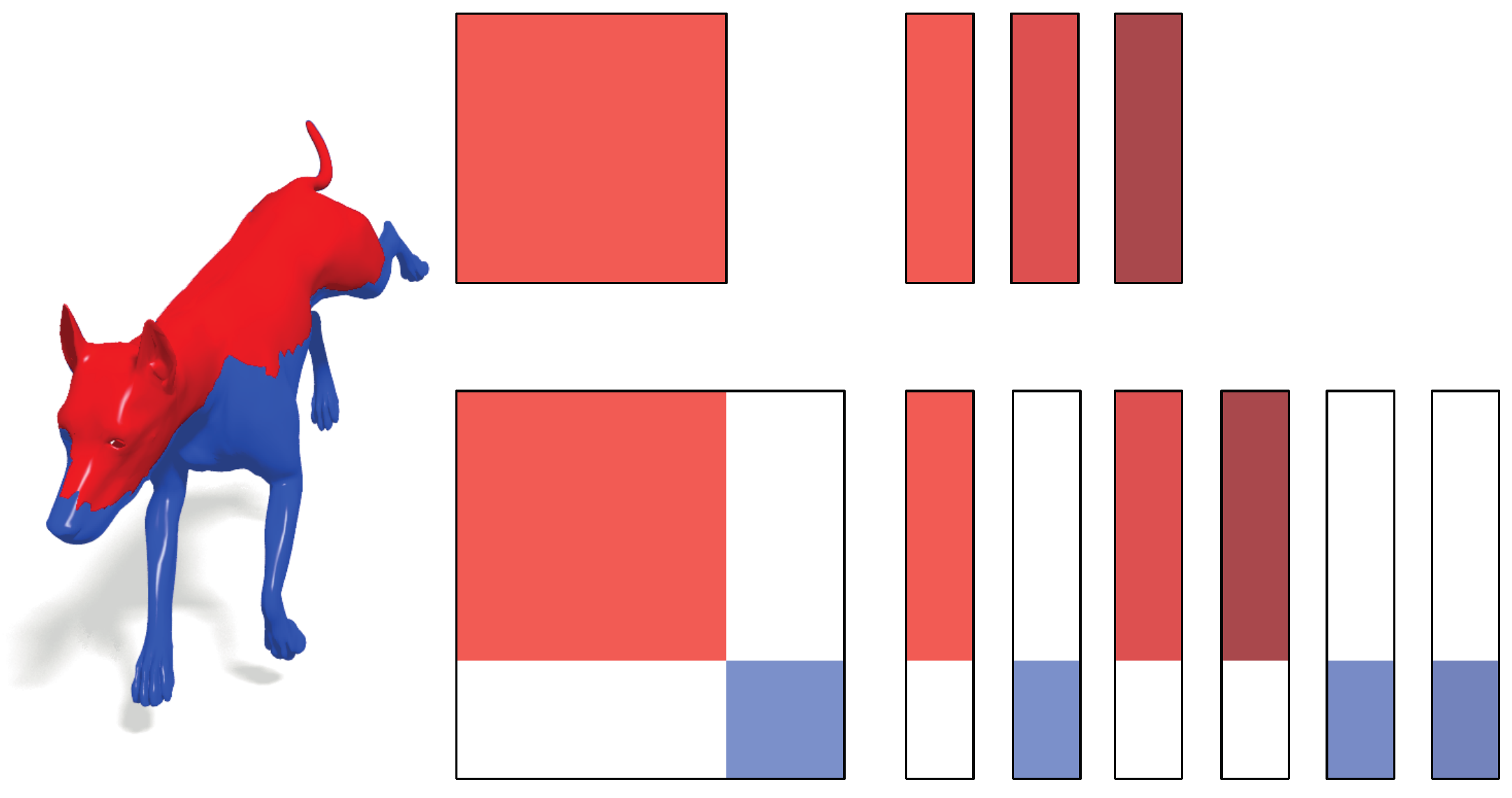}
	\put(36,43){\footnotesize $\mathbf{L}_\mathcal{N}$}
	\put(36,17){\footnotesize $\mathbf{L}_\mathcal{N}$}
	\put(49,4){\footnotesize $\mathbf{L}_{\overline{\mathcal{N}}}$ }
	\put(60.5,17){\footnotesize $\bm{\phi}_1$}	
	\put(74.5,17){\footnotesize $\bm{\phi}_2$}	
	\put(81.5,17){\footnotesize $\bm{\phi}_3$}	
	\put(60.5,42){\footnotesize $\bm{\phi}_1$}	
	\put(67.5,42){\footnotesize $\bm{\phi}_2$}	
	\put(74.5,42){\footnotesize $\bm{\phi}_3$}	
	\put(67.75,4){\footnotesize $\bar{\bm{\phi}}_1$}	
	\put(88.5,4){\footnotesize $\bar{\bm{\phi}}_2$}	
	\put(95.5,4){\footnotesize $\bar{\bm{\phi}}_3$}			
	\put(15,32){\footnotesize $\mathcal{N}$}	
	\put(14.5,23.5){\color{white}\footnotesize $\overline{\mathcal{N}}$}		
		\end{overpic}\\
  \caption{
  \label{fig:blocks}The eigenvalues and eigenvectors of a block-diagonal Laplacian $\mathbf{L}_{\mathcal{M}}$ are an interleaved sequence of the eigenpairs from the two blocks $\mathbf{L}_{\mathcal{N}}$ and $\mathbf{L}_{\overline{\mathcal{N}}}$. 
  }
\end{figure}

\section{Laplacian eigenvectors and eigenvalues under partiality}
\label{sec:perturb}


When one of the two shapes has missing parts, the assumption of approximate isometry does not hold anymore and a direct application of the method of Ovsjanikov~\etal~(\ie, solving system \eqref{eq:cab}) would not produce meaningful results. 
However, as we show in this section, the matrix $\vct{C}$ still exhibits a particular structure which can be exploited to drive the matching process.

We assume to be given a full shape $\mathcal{M}$ and a part thereof $\mathcal{N}\subset \mathcal{M}$.
We further denote by $\overline{\mathcal{N}} = \mathcal{M}\setminus \mathcal{N}$ the remaining vertices of $\mathcal{M}$. 
The manifolds $\mathcal{M}$ and $\mathcal{N}$ are discretized as triangular meshes with $m$ and $n$ vertices, respectively, and $\bar{n} = m-n$. 
The scenario we consider in this paper concerns the problem of matching an approximately isometric deformation of part $\mathcal{N}$ to the full shape $\mathcal{M}$ (part-to-whole matching). 
Our goal is to characterize the eigenvalues and eigenvectors of the Laplacian $\vct{L}_\mathcal{M}$ in terms of perturbations of the eigenvalues and eigenvectors of the Laplacians $\vct{L}_\mathcal{N}$ and $\vct{L}_{\overline{\mathcal{N}}}$~\cite{NME:NME1620260202}. We tacitly assume that homogeneous Neumann boundary conditions~(\ref{eq:neumann}) apply.

\subsection{Block-diagonal case}\label{sec:blockd}
%
%

For the simplicity of analysis, let us first consider a simplified scenario in which $\mathcal{N}$ and $\overline{\mathcal{N}}$ are {\em disconnected}, \ie, there exist no links between the respective boundaries $\partial \mathcal{N}$ and $\partial \overline{\mathcal{N}}$. 
W.l.o.g., we can assume that the vertices in $\mathcal{M}$ are ordered such that the vertices in $\mathcal{N}$ come before those in $\overline{\mathcal{N}}$. 
With this ordering, the $m\times m$ Laplacian matrix $\vct{L}_\mathcal{M}$ is block-diagonal, with an $n\times n$ block $\vct{L}_\mathcal{N}$ and an $\bar{n}\times \bar{n}$ block $\vct{L}_{\overline{\mathcal{N}}}$. 
%
The (sorted) eigenvalues of $\vct{L}_\mathcal{M}$ form a mixed sequence composed of the eigenvalues from $\vct{L}_\mathcal{N}$ and $\vct{L}_{\overline{\mathcal{N}}}$. Similarly, the eigenvectors of $\vct{L}_\mathcal{M}$ correspond to the eigenvectors of the two sub-matrices, zero-padded to the correct size (Fig.~\ref{fig:blocks}). 

\begin{figure*}[t]
  \centering
  \begin{overpic}
  [trim=0cm 0cm 0cm 0cm,clip,width=1\linewidth]{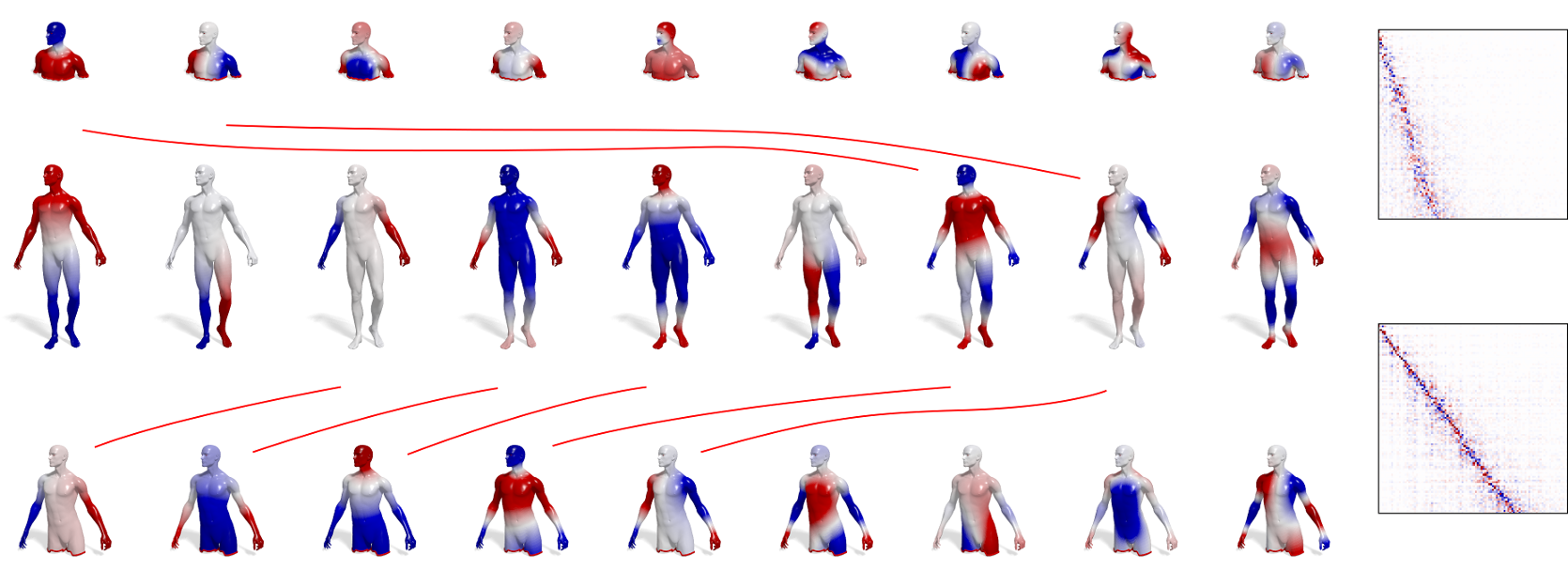}
  \put(3.5,0.1){\footnotesize $\phi_2$}
  \put(12.8,0.1){\footnotesize $\phi_3$}
  \put(22.5,0.1){\footnotesize $\phi_4$}
  \put(32.3,0.1){\footnotesize $\phi_5$}
  \put(42.1,0.1){\footnotesize $\phi_6$}
  \put(51.8,0.1){\footnotesize $\phi_7$}
  \put(61.5,0.1){\footnotesize $\phi_8$}
  \put(71.0,0.1){\footnotesize $\phi_9$}
  \put(80.9,0.1){\footnotesize $\phi_{10}$}
  \put(3.5,13){\footnotesize $\psi_2$}
  \put(12.8,13){\footnotesize $\psi_3$}
  \put(22.5,13){\footnotesize $\psi_4$}
  \put(32.3,13){\footnotesize $\psi_5$}
  \put(42.1,13){\footnotesize $\psi_6$}
  \put(51.8,13){\footnotesize $\psi_7$}
  \put(61.5,13){\footnotesize $\psi_8$}
  \put(71.0,13){\footnotesize $\psi_9$}
  \put(80.9,13){\footnotesize $\psi_{10}$}
  \put(3.5,30){\footnotesize $\zeta_2$}
  \put(12.8,30){\footnotesize $\zeta_3$}
  \put(22.5,30){\footnotesize $\zeta_4$}
  \put(32.3,30){\footnotesize $\zeta_5$}
  \put(42.1,30){\footnotesize $\zeta_6$}
  \put(51.8,30){\footnotesize $\zeta_7$}
  \put(61.5,30){\footnotesize $\zeta_8$}
  \put(71.0,30){\footnotesize $\zeta_9$}
  \put(80.9,30){\footnotesize $\zeta_{10}$}
  \put(0,35.5){\footnotesize $\mathcal{N}_1$}
  \put(0,26){\footnotesize $\mathcal{M}$}
  \put(0,8){\footnotesize $\mathcal{N}_2$}
  \put(90.5,2){\footnotesize $\langle \psi_i , T \phi_j \rangle$}
  \put(90.5,21){\footnotesize $\langle \psi_i, T\zeta_j\rangle$}
  \end{overpic}
  \caption{\label{fig:eigenfunctions}First ten eigenfunctions of a full shape $\mathcal{M}$ and two parts $\mathcal{N}_1,\mathcal{N}_2\subset\mathcal{M}$ with different surface area. All eigenfunctions of the partial shapes have a corresponding eigenfunction $\psi_i$ on the full shape for some $i$; the correspondence between eigenfunctions follows from the correspondence between eigenvalues (see also Fig. \ref{fig:spectra}). This is reflected in functional maps with different diagonal slopes, where the slope depends on the area ratios of the two surfaces (by Weyl's law).}
\end{figure*}

\paragraph*{Structure of C under partiality.}
Suppose we are given the first $k$ Laplace-Beltrami eigenvalues of the full shape $\mathcal{M}$ and of its part $\mathcal{N}$. Since the spectrum of $\mathbf{L}_\mathcal{M}$ is an interleaved sequence of the eigenvalues of $\mathbf{L}_\mathcal{N}$ and $\mathbf{L}_{\overline{\mathcal{N}}}$, only the first ${r < k}$ eigenvalues of $\mathbf{L}_\mathcal{N}$ will appear among the first $k$ eigenvalues of $\mathbf{L}_\mathcal{M}$. The remaining ${k-r}$ eigenvalues of $\mathbf{L}_\mathcal{N}$ will only appear further along the spectrum of $\mathbf{L}_\mathcal{M}$ (see Fig. \ref{fig:spectra} for an example where $k=50$ and $r=21$). 
The same argument holds for the associated eigenfunctions, as illustrated in Fig.~\ref{fig:eigenfunctions}: 
%
if $\bm{\phi}_i$ is an eigenvector of $\mathbf{L}_{\mathcal{N}}$, then $\mathbf{L}_{\mathcal{M}}$ also has an eigenvector $\bm{\psi}_j$ such that $\bm{\phi}_{i} = \mathbf{T} \bm{\psi}_j$, where $\mathbf{T} = (\mathbf{I}_{n\times n}, \, \mathbf{0})^\top$ and $i < j$.

This analysis leads us to the following simple observation: the partial functional map between $\mathcal{N}$ and $\mathcal{M}$ is represented in the spectral domain by the matrix of inner products $c_{ij} = \langle \mathbf{T}\bm{\phi}_i, \bm{\psi}_j \rangle_{\mathcal{M}}$, which has a {\em slanted-diagonal} structure with a slope $r/k$ (see examples in Figs.~\ref{fig:teaser}, \ref{fig:eigenfunctions} where this structure is manifested approximately). 
Consequently, the last $k-r$ columns of matrix $\mathbf{C}$ are zero such that $r = \mathrm{rank}(\vct{C})$.  The value $r$ can be estimated by simply comparing the spectra of the two shapes, as shown in Fig.~\ref{fig:spectra}.
Note that this behavior 
is consistent with Weyl's asymptotic law \cite{weyl11}, according to which the Laplacian eigenvalues grow linearly, with rate inversely proportional to surface area. 

%
%


%

\subsection{Perturbation analysis}

We will now show that these properties still approximately hold when the Laplacian matrix $\mathbf{L}_\mathcal{M}$ is not perfectly block-diagonal, \ie, when 
%
$\mathcal{N}$ and $\overline{\mathcal{N}}$ are joined along their boundaries. 
Roughly speaking, the main observation is that in this case as well the matrix $\mathbf{C}$ has a slanted diagonal structure, where the diagonal angle depends on the relative area of the part, and the diagonal `sharpness' depends on the position and length of the cut.

\begin{figure}[b]
  \centering
		\begin{overpic}
		[trim=0cm 0cm 0cm 0cm,clip,width=0.725\linewidth]{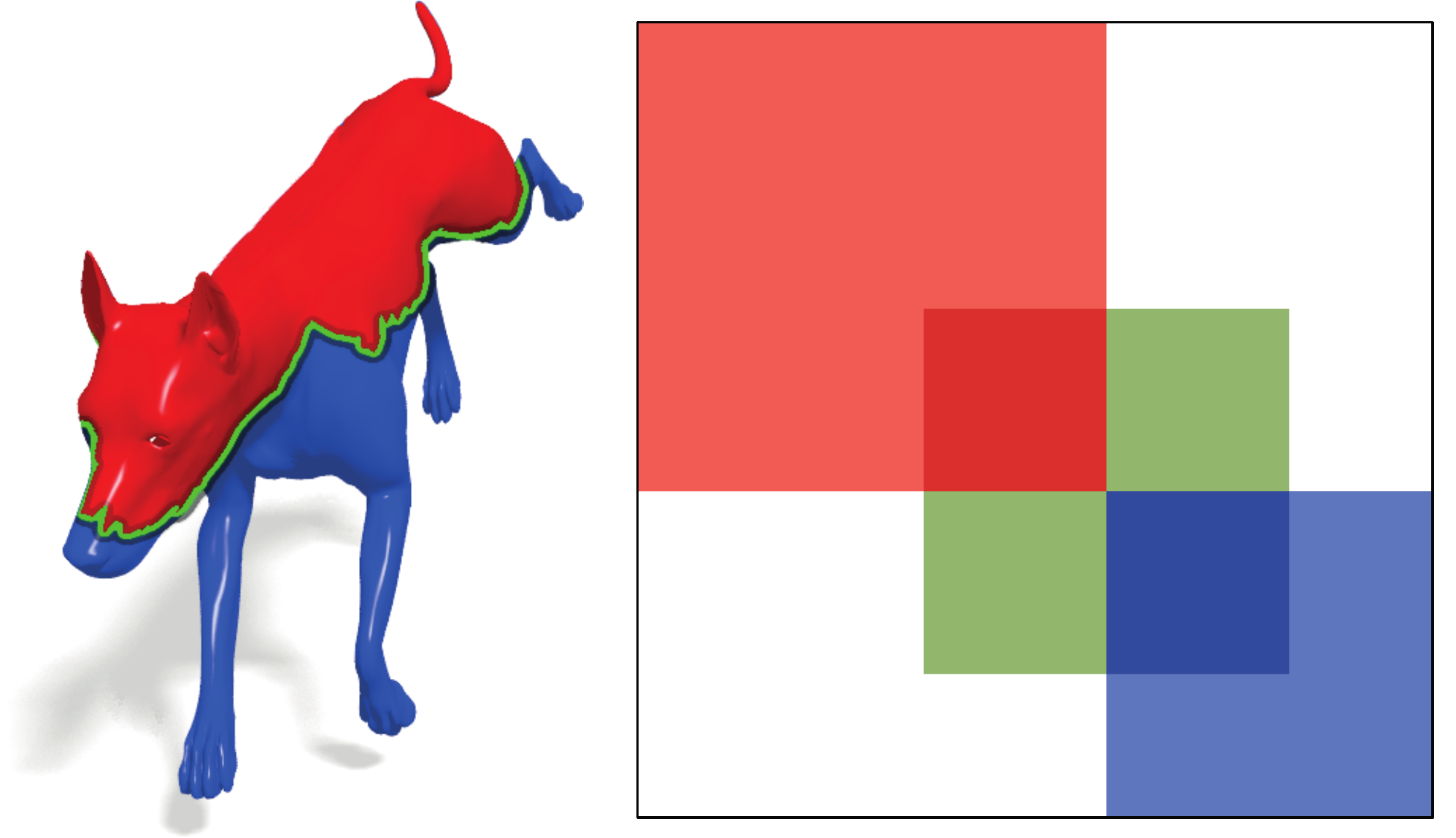}
	\put(52,44){\footnotesize $\mathbf{L}_\mathcal{N}$}
	\put(89,5){\footnotesize $\mathbf{L}_{\overline{\mathcal{N}}}$ }
%
%
	\put(79.5,28){\footnotesize $t\mathbf{E}$}
	\put(67,16.5){\footnotesize $t\mathbf{E}^\top$}
	\put(20,38.5){\footnotesize $\mathcal{N}$}	
	\put(20.25,27){\color{white}\footnotesize $\overline{\mathcal{N}}$}		
		\end{overpic}\\
  \caption{
  \label{fig:blocks2} The matrix $\mathbf{L}(t)$ is obtained as a perturbation of the block-diagonal Laplacian 
  in the boundary band (shown in green).   
  }
\end{figure}

Here we assume w.l.o.g. that within $\mathcal{N}$ the boundary vertices $\partial \mathcal{N}$ are indexed at the end, while within $\overline{\mathcal{N}}$ the boundary vertices $\partial \overline{\mathcal{N}}$ are indexed at the beginning. 
Then, there is a boundary band $\mathcal{B}=\partial \mathcal{N} \cup \partial \overline{\mathcal{N}}$ 
 such that only the entries of the Laplacians $\vct{L}_\mathcal{N}$ and $\vct{L}_{\overline{\mathcal{N}}}$ between vertices in $\mathcal{B}$ are affected by the cut (Fig.~\ref{fig:blocks2}).

We define the parametric matrix 
\begin{equation}
 \vct{L}(t) = 
 \left(\begin{array}{c | c}
 \vct{L}_\mathcal{N} & \mathbf{0}\\\hline 
 \mathbf{0} & \vspacer{10pt}\vct{L}_{\overline{\mathcal{N}}}
 \end{array}\right) + 
   t \left(\begin{array}{c | c}
  \vct{P}_\mathcal{N} & \vct{P} \\\hline
  \vct{P}^\top &\vspacer{10pt} \vct{P}_{\overline{\mathcal{N}}}
  \end{array}\right),
\end{equation}
where 
\begin{eqnarray*}
\vct{P}_\mathcal{N} =  \left(\begin{array}{c c}
 \mathbf{0} & \mathbf{0}\\
 \mathbf{0} & \vct{D}_{\mathcal{N}}
 \end{array}\right),  \,\,\,
\vct{P}_{\overline{\mathcal{N}}} =  \left(\begin{array}{c c}
 \vct{D}_{\overline{\mathcal{N}}} & \mathbf{0}\\
 \mathbf{0} & \mathbf{0}
 \end{array}\right),   \,\,\,
\vct{P} =  \left(\begin{array}{c c}
 \mathbf{0} & \mathbf{0}\\
 \vct{E}  & \mathbf{0}
 \end{array}\right)
\end{eqnarray*}
are matrices of size $n\times n$, $\bar{n}\times \bar{n}$, and $n\times \bar{n}$, respectively. 
%
%
Here $\vct{D}_\mathcal{N}$ and $\vct{D}_{\overline{\mathcal{N}}}$ represent the variations of the Laplacians $\vct{L}_\mathcal{N}$ and $\vct{L}_{\overline{\mathcal{N}}}$ {\em within} nodes in $\partial \mathcal{N}$ and $\partial \overline{\mathcal{N}}$ respectively, while $\vct{E}$ represents the variations {\em across} the boundary. 
The parameter $t$ is such that $\vct{L}(1)=\vct{L}_\mathcal{M}$, while for ${t=0}$ we get back to the disconnected case of Fig.~\ref{fig:blocks}. 
\rev{In what follows, we perform a differential analysis at ${t=0}$ and therefore analyze the change in eigenvalues and eigenvectors of $\vct{L}(t)$ as we interpolate between the seen ($\mathcal{N}$) and unseen ($\overline{\mathcal{N}}$) parts for $t=0$, to the full shape $\mathcal{M}$ for $t=1$.}

Note that with the appropriate ordering of the vertices, 
 the matrices $\vct{D}_\mathcal{N}$ and $\vct{D}_{\overline{\mathcal{N}}}$ will have a band-diagonal structure.
%
In fact, a cut through an edge will affect the values of the discrete Laplacian matrix $\mathbf{L}$ only at the entries corresponding to the vertices at the extremities of the edge, and to the edges laying in the same triangle as the cut edge. For example, looking at Fig.~\ref{fig:cot_weights}, a cut through edge $(i,j)$ will affect the diagonal entries $l_{ii}$ and  $l_{jj}$ as well as the off-diagonal entries $l_{ih}$, $l_{ik}$, $l_{jh}$, and $l_{jk}$. Note also that the continuity of the cut implies that two of the four off-diagonal entries will be cut as well, leaving  no more than two affected edges on any side of the cut. 
As a result, the entries of the Laplacian affected by the cut correspond to the nodes and edges in a path along the boundary of the cut.

\rev{We further note that although here we consider the cotangent Laplacian for simplicity of analysis, similar results hold for Laplacians that are not strictly local, but locally dominant \cite{CLOPC_SODA_09}, \ie, most of their $L_2$ norm is due to the elements in a tight boundary layer.}

\newtheorem{theorem}{Theorem}
\begin{theorem}\label{thm:evals}
Let $\vct{L}_\mathcal{N}+t\vct{P}_\mathcal{N} = \bm{\Phi}(t)\T \bm{\Lambda}(t) \bm{\Phi}(t)$, where $\bm{\Lambda}(t) = \mathrm{diag}(\lambda_1(t), \hdots, \lambda_n(t))$ is a diagonal matrix of eigenvalues, and $\bm{\Phi}(t)$ are the corresponding eigenvectors. 
The derivative of the non-trivial eigenvalues is given by 
\begin{equation}\label{eq:eigdt}
 \frac{d}{dt}\lambda_i = \sum_{v,w\in \partial \mathcal{N}} (\vct{P}_\mathcal{N})_{vw} {\phi}_{iv} {\phi}_{iw}
 = \bm{\phi}_i^\top \vct{P}_\mathcal{N} \bm{\phi}_i. 
\end{equation}
\end{theorem}
{\em Proof:} See Appendix C.

\begin{figure}[t]
  \centering
\setlength\figureheight{3.25cm} 
\setlength\figurewidth{0.6\columnwidth}    
%
%
\begin{tikzpicture}

\begin{axis}[%
width=0.95092\figurewidth,
height=\figureheight,
at={(0\figurewidth,0\figureheight)},
scale only axis,
every outer x axis line/.append style={black},
every x tick label/.append style={font=\color{black}, font=\footnotesize},
xmin=0,
xmax=50,
xlabel={\footnotesize eigenvalue number},
xlabel near ticks,
xtick={10,20,30,40,50},
every outer y axis line/.append style={black},
every y tick label/.append style={font=\color{black}, font=\footnotesize, /pgf/number format/.cd,
            fixed,
            fixed zerofill,
            precision=2,
        /tikz/.cd},
ymin=0,
ymax=0.08,
axis background/.style={fill=white},
axis x line*=bottom,
axis y line*=left,
]
\addplot [color=white!30!black,dotted,line width=1.2pt,forget plot]
  table[row sep=crcr]{%
21	0.0212649650692466\\
22	0.0212649650692466\\
23	0.0212649650692466\\
24	0.0212649650692466\\
25	0.0212649650692466\\
26	0.0212649650692466\\
27	0.0212649650692466\\
28	0.0212649650692466\\
29	0.0212649650692466\\
30	0.0212649650692466\\
31	0.0212649650692466\\
32	0.0212649650692466\\
33	0.0212649650692466\\
34	0.0212649650692466\\
35	0.0212649650692466\\
36	0.0212649650692466\\
37	0.0212649650692466\\
38	0.0212649650692466\\
39	0.0212649650692466\\
40	0.0212649650692466\\
41	0.0212649650692466\\
42	0.0212649650692466\\
43	0.0212649650692466\\
44	0.0212649650692466\\
45	0.0212649650692466\\
46	0.0212649650692466\\
47	0.0212649650692466\\
48	0.0212649650692466\\
49	0.0212649650692466\\
50	0.0212649650692466\\
};
\addplot [color=white!30!black,dotted,line width=1.2pt,forget plot]
  table[row sep=crcr]{%
21	0\\
21	0.00106324825346233\\
21	0.00212649650692466\\
21	0.00318974476038699\\
21	0.00425299301384932\\
21	0.00531624126731165\\
21	0.00637948952077397\\
21	0.0074427377742363\\
21	0.00850598602769863\\
21	0.00956923428116096\\
21	0.0106324825346233\\
21	0.0116957307880856\\
21	0.0127589790415479\\
21	0.0138222272950103\\
21	0.0148854755484726\\
21	0.0159487238019349\\
21	0.0170119720553973\\
21	0.0180752203088596\\
21	0.0191384685623219\\
21	0.0202017168157843\\
21	0.0212649650692466\\
};
\addplot [color=blue,line width=0.5pt,mark size=0.5pt,only marks,mark=ball,mark options={solid}]
  table[row sep=crcr]{%
1	7.85368948694187e-018\\
2	0.000203466733902504\\
3	0.000421910448365916\\
4	0.000522474785278309\\
5	0.00059090838558119\\
6	0.00074664832700896\\
7	0.00104654276032573\\
8	0.00142032825520474\\
9	0.00224112813931745\\
10	0.00252974341050051\\
11	0.00261187856761526\\
12	0.0030647981202222\\
13	0.00361093737351105\\
14	0.0037751440419233\\
15	0.00465775283349989\\
16	0.00515772437169289\\
17	0.00561751455245374\\
18	0.00589096696162439\\
19	0.00613019523575216\\
20	0.00655116328232825\\
21	0.00702083961417513\\
22	0.00759464289166461\\
23	0.00792160690543119\\
24	0.00845374494194468\\
25	0.0086630687632443\\
26	0.00939503747883993\\
27	0.00990508698474723\\
28	0.0102475919140669\\
29	0.010509323879634\\
30	0.0106665249560853\\
31	0.0108655977518782\\
32	0.011581260394464\\
33	0.012228700045812\\
34	0.0130998878205399\\
35	0.0134408951056184\\
36	0.0138517494847631\\
37	0.0144643371501675\\
38	0.0149046074555975\\
39	0.0153851039990854\\
40	0.0159431571173271\\
41	0.0165087268183912\\
42	0.0168987248860711\\
43	0.0170423591042677\\
44	0.0178037440514079\\
45	0.0185457098598966\\
46	0.0191149360252548\\
47	0.0199326091230905\\
48	0.0204886876501593\\
49	0.0208695124519082\\
50	0.0213305808747694\\
};

\addplot [color=red,line width=0.5pt,mark size=0.5pt,only marks,mark=ball,mark options={solid}]
  table[row sep=crcr]{%
1	7.75882179684939e-019\\
2	0.000198859369454982\\
3	0.000927168937211446\\
4	0.00136898231088931\\
5	0.00170874659198827\\
6	0.00379914580350586\\
7	0.00442915228231305\\
8	0.005269429442467\\
9	0.00616363965752829\\
10	0.00812198066661168\\
11	0.00855883526283856\\
12	0.00930006403752559\\
13	0.0106070243803978\\
14	0.0127815653794653\\
15	0.0133143005083636\\
16	0.0137522143934158\\
17	0.0156570060213649\\
18	0.0178626434962966\\
19	0.0183752944944034\\
20	0.0196284784844719\\
21	0.0212649650692466\\
22	0.0227568244743238\\
23	0.0233768155214945\\
24	0.0244541856717368\\
25	0.0279605519156909\\
26	0.0288630381133013\\
27	0.0303614255866276\\
28	0.0324438277004751\\
29	0.0335131312114702\\
30	0.03467510249967\\
31	0.0356663039209743\\
32	0.0364886142340272\\
33	0.038835939688361\\
34	0.0392676307097299\\
35	0.0402867187136446\\
36	0.0419657964765039\\
37	0.0425821127365483\\
38	0.0437754237256062\\
39	0.0464665754851196\\
40	0.0479540680055988\\
41	0.0493060311119224\\
42	0.0519052537066578\\
43	0.0529188345775512\\
44	0.0539391115408294\\
45	0.0554278725765621\\
46	0.0571906024999261\\
47	0.0579959187632752\\
48	0.059122831022708\\
49	0.0613877239283652\\
50	0.0626619408597348\\
};

\addplot [color=red,only marks,mark=o,mark options={solid},forget plot]
  table[row sep=crcr]{%
21	0.0212649650692466\\
};
\addplot [color=blue,only marks,mark=o,mark options={solid},forget plot]
  table[row sep=crcr]{%
50	0.0213305808747694\\
};
\end{axis}
\end{tikzpicture}%
    \includegraphics[width=0.24\linewidth]{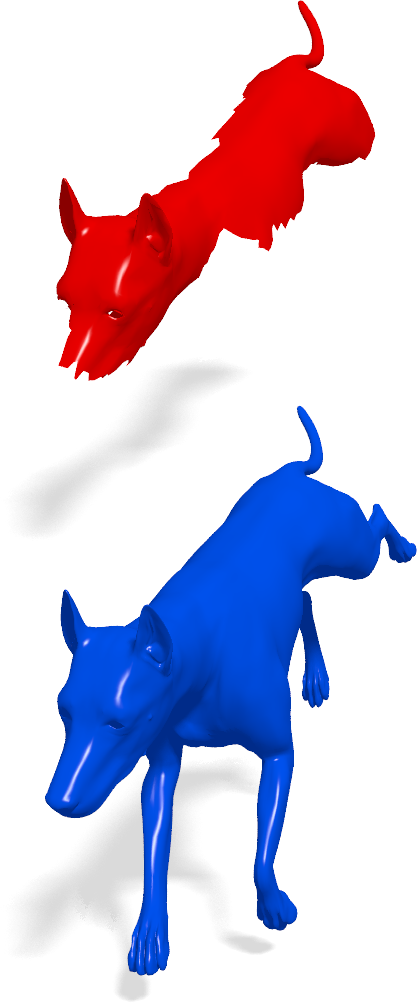}
  \caption{\label{fig:spectra}Neumann spectra of a full shape and a part of it. The eigenvalues of the partial shape (in red) are approximately preserved under the partiality transformation (see Theorem~\ref{thm:evals}), and appear perturbed in the spectrum of the full shape (in blue). This simple observation allows us to estimate the diagonal slope of the functional map relating the two shapes; in this example, the slope is equal to $21/50$.}
\end{figure}

Theorem~\ref{thm:evals} establishes that the (first-order) change in the eigenvalues of the partial shape $\mathcal{N}$ only depends on the change in the
Dirichlet energy of the corresponding eigenvectors along the boundary $\partial\mathcal{N}$ 
%
 \rev{(recall that for eigenvector $\bm{\phi}_i$, the Dirichlet energy is defined as $\bm{\phi}_i\T (\vct{L}_\mathcal{N}+t\vct{P}_\mathcal{N})\bm{\phi}_i = \bm{\phi}_i\T \vct{L}_\mathcal{N} \bm{\phi}_i + t\bm{\phi}_i\T \vct{P}_\mathcal{N} \bm{\phi}_i$).}
 This means that the eigenvalues are perturbed depending on the {\em length} and {\em position} of the cut. 
 \rev{Note that the estimate given in Eq.~\eqref{eq:eigdt} can not typically be computed directly, as this would assume knowledge of the correspondence between $\mathcal{N}$ and $\mathcal{M}$. However,} by virtue of this result, we can establish approximate correspondence between the eigenvalues \rev{$\lambda_i^\mathcal{N}$} of $\mathbf{L}_\mathcal{N}$ and a subset of the eigenvalues \rev{$\lambda_j^\mathcal{M}$} of $\mathbf{L}_\mathcal{M}$ (which are now not exactly equal as in the block-diagonal case). 
%
\rev{We do this in order to estimate the slope of $\vct{C}$. Specifically, we compute
\begin{equation}
r = \max \{ i ~|~ \lambda_i^\mathcal{N} < \max_{j=1}^k \lambda_j^\mathcal{M} \}\,;
\end{equation}
the slope of $\vct{C}$ can now be estimated as $r/k$, as explained in Section \ref{sec:blockd} (see also Fig.~\ref{fig:spectra} for a visual illustration of the estimation of $r$).
}


\begin{theorem}\label{thm:evecs} 
Assume that $\vct{L}_{{\mathcal{N}}}$ has distinct eigenvalues ($\lambda_i \neq \lambda_j$ for $i\neq j$), and furthermore, the non-zero eigenvalues are all distinct from the eigenvalues of $\vct{L}_{\overline{\mathcal{N}}}$ ($\lambda_i \neq \overline{\lambda}_j$ for all $i, j$). 
Let $\vct{L}_\mathcal{N}+t\vct{P}_\mathcal{N} = \bm{\Phi}(t)\T \bm{\Lambda}(t) \bm{\Phi}(t)$, where $\bm{\Lambda}(t) = \mathrm{diag}(\lambda_1(t), \hdots, \lambda_n(t))$ is a diagonal matrix of eigenvalues, and $\bm{\Phi}(t)$ are the corresponding eigenvectors.
Then, the derivative of the non-constant eigenvector 
is given by 
\begin{equation}
 \frac{d}{dt}\bm{\phi}_i = \sum_{ {j=1}\atop{j\neq i}}^{n}  \frac{\bm{\phi}_i\T \vct{P}_\mathcal{N} \bm{\phi}_j}{{\lambda}_i-{\lambda}_j} \bm{\phi}_j
 + \sum_{j=1}^{\overline{n}}  \frac{\bm{\phi}_i\T \vct{P}\; \overline{\bm{\phi}}_j}{{\lambda}_i-\overline{{\lambda}}_j} \overline{\bm{\phi}}_j\,.
 \label{eq:thrm2}
\end{equation}
\end{theorem}
{\em Proof:} See Appendix C.

\paragraph*{Remark.} 
If $\vct{L}_{\overline{\mathcal{N}}}$ shares some eigenvalues with $\vct{L}_{\mathcal{N}}$, the second sum in~(\ref{eq:thrm2}) would be slightly different~\cite{NME:NME1620260202}, but would still only have support over  $\overline{\mathcal{N}}$.

We conclude from Theorem~\ref{thm:evecs} that the perturbation associated with the partiality transformation gives rise to a mixing of eigenspaces. 
%
%
The second summation in~(\ref{eq:thrm2}) has support over $\overline{\mathcal{N}}$ and thus provides the completion of the eigenfunction on the missing part. 
The first summation in~(\ref{eq:thrm2}) is responsible for the modifications of the eigenvectors over the nodes in $\mathcal{N}$. 
Here the numerator has a term $\bm{\phi}_i\T \vct{P}_\mathcal{N} \bm{\phi}_j$ which, since $\vct{D}_\mathcal{N}$ is band-diagonal and diagonally dominant, acts as a dot product of the eigenvectors over the boundary band. This points to large mixing of eigenvectors with a strong co-presence near the boundary. In turn, the term ${\lambda}_i-{\lambda}_j$ at the denominator forces a strong mixing of eigenvectors corresponding to similar eigenvalues.
This results in an amplification of the variation for higher eigenvalues, as eigenvalues tend to densify on the higher end of the spectrum, and explains the funnel-shaped spread of the matrix $\mathbf{C}$ visible at high frequencies (see Fig.~\ref{fig:C}).
%

%

\begin{figure}[t]
  \centering
  \begin{overpic}
		[trim=0cm 0cm 0cm 0cm,clip,width=1\linewidth]{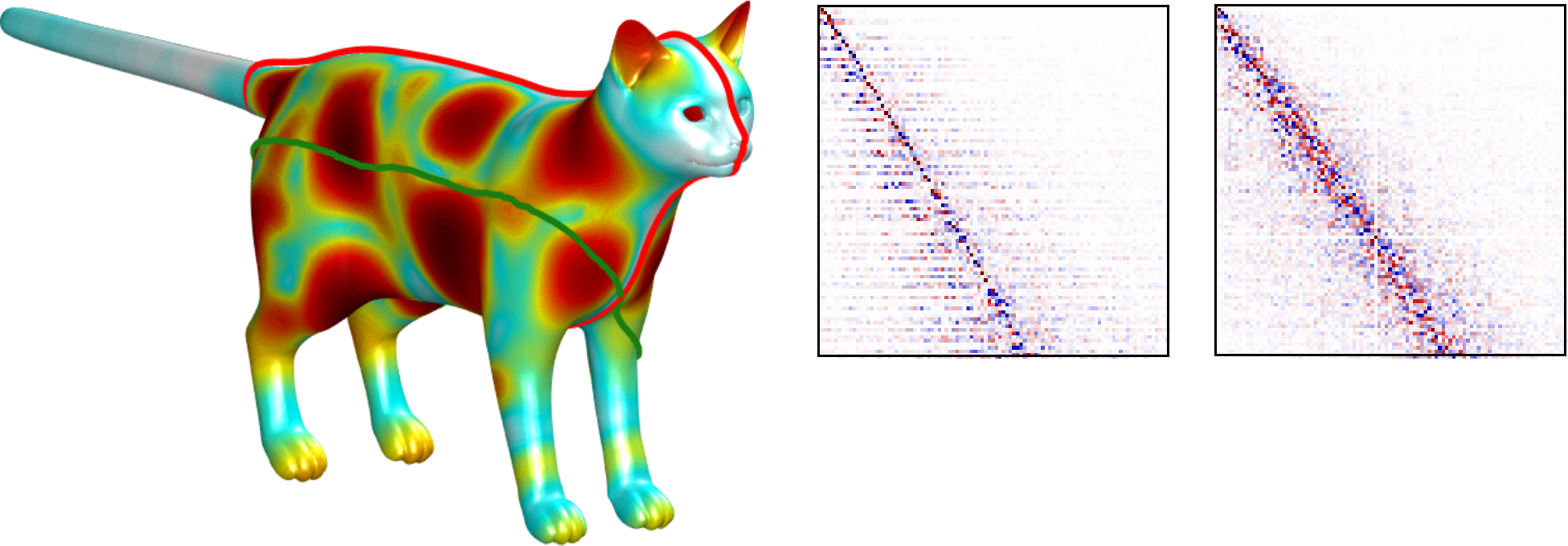}
  \put(49,0){\input{f1.tikz}}
  \put(77,0){
%
%
\definecolor{mycolor1}{rgb}{0.10588,0.49020,0.05882}%
\begin{tikzpicture}

\begin{axis}[%
width=0.22\linewidth,
height=0.10\linewidth,
scale only axis,
separate axis lines,
every outer x axis line/.append style={black},
every x tick label/.append style={font=\color{black}, font=\tiny},
xmin=0,
xmax=300,
every outer y axis line/.append style={black},
every y tick label/.append style={font=\color{black}, font=\tiny},
yticklabel=\empty,
xticklabel=\empty,
xlabel near ticks,
ymin=0,
ymax=25,
axis background/.style={fill=white},
inner sep=0pt
]
\addplot [color=mycolor1,solid,line width=0.9pt,forget plot]
  table[row sep=crcr]{%
0	2.07509489565895\\
0.714970277431168	2.11809669287291\\
1.50770987712075	2.26809929704248\\
4.01794342125604	1.70948752500289\\
5.55589346218715	2.14063246590778\\
7.34296690533435	1.92364848553458\\
8.86997830015169	2.55195861963575\\
10.0862961750109	3.5159484938942\\
11.1338494853966	4.36322000472053\\
12.8133598538076	7.21755364926043\\
13.8528974668375	8.16190771408473\\
15.0437512575709	8.95996783385012\\
16.8346789764093	11.8918994134954\\
18.3166714387397	12.0583519009688\\
19.9273863987509	11.3211613508696\\
22.2299590872029	11.7130149068526\\
24.101240885116	8.36404316893598\\
25.9823025924518	4.65614548528613\\
28.273732165626	3.44801431959203\\
30.0699616043136	5.28539571095046\\
31.7999130794137	7.40790581487962\\
33.4249806878546	8.11654959925378\\
34.6986902891394	8.01314727992859\\
36.9355009196596	6.02990984025157\\
39.1384055599034	3.80908324534807\\
41.4057190733817	5.39656965077169\\
42.635626105641	7.2630699880149\\
45.0181526529021	11.3371402550672\\
46.1854387182696	13.6793297391239\\
48.0692886073544	14.4451921637648\\
49.2454769079271	16.6520510262497\\
50.6477015906399	19.0142006697096\\
52.3780934861968	20.8750272601997\\
54.072143228797	18.9586256384066\\
56.0383983135205	18.6208342515368\\
58.1904065084166	15.4606891733177\\
60.4268010267208	12.2410840915294\\
62.5691605960996	8.10646695219622\\
64.5368895823671	5.49710765064693\\
66.466636510339	4.61205030222023\\
68.5942460043488	5.7617576003278\\
70.0567380850454	6.05031071216018\\
72.211429380707	6.33388203066258\\
74.3669867980819	5.70694459595843\\
76.5952879670678	5.34135808809472\\
79.1899932999719	3.1027819103065\\
81.7366971570391	4.96477627366264\\
84.3322252322103	7.28754518372659\\
86.6391874807099	8.83357524825811\\
88.3739111656822	9.09848879288781\\
89.6957201217738	8.66556812302272\\
90.8289972179869	7.9715319892213\\
91.8233864196584	7.25662210991762\\
92.6903930747655	6.59204321648336\\
93.4608092297117	6.03242400068835\\
94.1801162302344	5.60561168108222\\
94.9516742519055	5.26232791928127\\
95.8720343707742	5.06578267752139\\
96.9470181347455	5.11771502371256\\
98.5263880375865	5.75993066813947\\
99.5756788842082	6.03044849377649\\
100.66694196404	6.28848887302451\\
101.769221597529	6.28943593005202\\
103.541797771748	6.84556712485107\\
104.612392451261	6.68841475892681\\
105.66246839966	6.52440114794641\\
107.541996130235	6.22580827077874\\
108.598961455212	5.9049373418786\\
109.647829272624	5.52061705680097\\
110.673423377914	5.10281281834421\\
112.4789025614	3.92975635732051\\
113.444919582679	3.3701512720417\\
114.356560882095	2.98471637910115\\
115.207130631856	2.78164825860062\\
115.952210026692	2.69610705009997\\
117.722910331488	4.35290533046968\\
118.493249371782	4.87502332447416\\
119.201895162578	5.81771006141457\\
119.877913560412	6.44705083410106\\
120.589181311828	6.89642082581363\\
123.372360760822	11.1175035296771\\
124.884794242269	12.3677203389417\\
126.482783838012	13.7621302204842\\
127.668260838592	14.000725868224\\
129.39766428892	14.1368795882543\\
130.54267979563	12.9833079526511\\
132.204548955673	13.1733777949844\\
133.452650840458	11.3223150185159\\
135.245341581445	10.6248798151915\\
136.412438478238	8.32919654530583\\
137.974012262088	7.39390403787654\\
139.03288591305	4.98220635380597\\
140.459268945269	4.20316718143857\\
141.870886153304	1.79109007924997\\
143.389415177003	0.389969192408071\\
144.106354512949	1.53845317605322\\
145.548900855587	1.46114488355689\\
146.294978652697	2.77063186069326\\
147.927027687261	2.89541216805709\\
149.342548448436	3.89653433613987\\
150.7761841206	3.33762187562651\\
151.879198834418	3.80616199380899\\
152.855549797298	3.9973249870219\\
153.770888606434	4.06911401972424\\
154.734987854698	4.0581760757379\\
155.77940856074	4.01199229171036\\
156.97540094479	3.99831993478978\\
158.340182430633	3.9611456214107\\
159.635566386069	3.97277477949634\\
160.325378342186	3.88031840232956\\
161.716793297044	3.84275907924561\\
162.383210804321	3.74341447736124\\
163.855231868234	3.41653343521618\\
164.575079798843	3.2861733339317\\
166.157210364568	2.81792937508531\\
167.080729573248	2.80551590507692\\
168.276759781482	3.15023065524791\\
169.569057108339	3.1210230036578\\
170.581186461345	3.77654352688083\\
171.639642941509	4.55062641311639\\
172.87427451927	5.427451953362\\
174.239026783237	6.3360383262582\\
175.769782898366	5.17425863038524\\
177.19085176468	5.8419553253125\\
178.365496976138	6.19051963163327\\
179.926650459944	4.21418113982828\\
180.967690563303	4.27219619236463\\
181.99314679962	4.56241113986153\\
183.509877563863	2.48969921994291\\
184.576550235024	2.77725406874359\\
185.93701095037	2.0251207307791\\
186.922603432081	1.73999109075065\\
188.158610323405	2.47465363822399\\
189.117890865114	1.7709467290577\\
190.014571989598	2.0517531505926\\
191.019774897723	2.16171834715197\\
191.592678943441	2.25976491164472\\
192.058192326602	2.28019899527768\\
193.149428236799	3.23786720893061\\
193.693676778919	3.32700109443832\\
194.312276999337	3.19664714611484\\
195.766489372171	4.50377814823856\\
196.740820530252	3.83293814913027\\
198.131276790946	4.85634956126603\\
199.646467778447	5.20955862221911\\
201.432885601942	6.14989180452027\\
203.439014602881	6.87729936933956\\
204.358953332472	6.03544570450897\\
206.570884497521	6.22974915825204\\
208.830658749676	5.66266670945806\\
209.853590443829	5.21245075022649\\
212.036323283597	4.40389213263735\\
214.0943544243	3.3300330256094\\
216.039123677579	2.31297913013441\\
217.95076774071	1.59397945374962\\
219.889397891383	1.16308446101883\\
221.734714061915	1.61485412455515\\
223.363133855753	3.03882967897831\\
224.855829304028	4.83910720646859\\
226.419079204524	7.27215188800732\\
227.963190113397	9.71150423948183\\
229.613699905704	12.2896098085444\\
231.20610014776	14.5223259966143\\
232.434042971633	16.710345315046\\
233.610515580791	17.9416201734625\\
234.773496595081	18.8870968298373\\
235.735663380884	20.1941025135276\\
236.74363900539	20.8287737425241\\
237.583830942689	21.2583539007851\\
238.281198349154	21.7671197198309\\
238.96537910758	21.6827793413548\\
239.83434171351	21.2211524262742\\
240.618588661419	21.6307158926325\\
241.704327423829	20.6959464949534\\
242.761725754639	19.3768849506421\\
244.036764224961	17.4303295946104\\
245.402507237025	15.3550229865899\\
246.853623248859	12.7034603079825\\
248.543576789878	9.71730035220415\\
250.181375818392	7.65406925239648\\
251.419668059505	6.69280489461584\\
252.204965142144	7.38189160289873\\
254.058720245334	7.81347900553491\\
256.099162695637	8.31497270101346\\
257.212589344501	8.81679370165883\\
258.714146863147	8.14502974149126\\
260.06987423412	8.49761566323349\\
261.603116257027	7.67738510319096\\
262.773190558086	7.55534557612173\\
264.455294085816	6.4026676066877\\
265.258466653826	5.85356968062844\\
265.988628996597	5.28632867072041\\
266.746299868415	4.70989049293943\\
267.975994789443	4.08272538653516\\
268.847402580206	3.33964266062364\\
270.131137467716	2.75855093741914\\
270.985869086417	2.07033278752036\\
272.157093385519	2.41614626231604\\
273.257522704649	3.2917084505458\\
274.174226428325	4.64820325972154\\
274.917081813384	4.12946589642551\\
275.832965895305	5.33562677401033\\
276.592021828226	4.70100478880915\\
277.555017335277	5.69080885690927\\
278.306992276682	4.98070988420125\\
279.337788969973	5.68577493346821\\
280.431100231784	6.17257128185151\\
281.243996830928	5.25932503305368\\
282.259508981381	5.32576935642083\\
283.234840779801	4.40468797944416\\
284.119053561036	4.24302448914603\\
284.949479293119	4.01763011628358\\
285.735606654181	3.82823293329015\\
286.519622727619	3.67257079136475\\
287.825840904421	3.00424753821197\\
288.474119476565	2.89536270616771\\
289.150716361941	2.77209560354961\\
291.356294205593	3.13895081520391\\
292.26104200377	2.91123013558846\\
293.20183553075	2.66694524175647\\
294.053791317636	2.48512001323682\\
294.732318292971	2.38652266554771\\
295.313819778927	2.35460659961666\\
};
\end{axis}
\end{tikzpicture}%
}
  \end{overpic}
  \caption{\label{fig:cuts}{\em Left}: A model is cut in two different ways (red and green curves) with cuts of same length. The off-diagonal dispersion depends mainly on the position of each cut. Function $f$ \eqref{eq:f} is plotted over the model. {\em Middle}: Ground-truth functional map between the complete model and the partial shape produced by the red cut (top), and values of $f$ along the cut (bottom). {\em Right}: Plots associated to the green cut.}
\end{figure}

Similarly to the case of eigenvalues, the eigenvectors are also perturbed depending on the length and position of the cut. 
The variation of the eigenvectors due to the mixing within the partial shape can be reduced  either by shortening the boundary of the cut, or by reducing the strength of the boundary interaction. The latter can be achieved by selecting a boundary along which eigenvectors with similar eigenvalues are either orthogonal, or both small. 
The {\em boundary interaction strength} can be quantified by considering the following function (we refer to Appendix C for a derivation):
\begin{equation}\label{eq:f}
f(v) =  \sum_{{i,j=1}\atop{j\neq i}}^{n} \left(\frac{{\phi}_{iv}{\phi}_{jv}}{{\lambda}_i-{\lambda}_j}\right)^2\,.
\end{equation}
Fig.~\ref{fig:cuts} shows an example of two different cuts with different interaction strengths, where the function $f$ is plotted on top of the cat model.
The cuts plotted in the figure have the same length, but one cut goes along a symmetry axis of the shape and through low values of $f$, while the other goes through rather high values of $f$. This is manifested in the dispersion of the slanted diagonal structure of the matrix $\mathbf{C}$ (larger in the second case). 


%

\section{Partial functional maps}
\label{sec:method}

%
%

As stated before, throughout the paper we consider the setting where we are given a full model shape $\mathcal{M}$ and another query shape $\mathcal{N}$ that corresponds to an approximately isometrically deformed part $\mathcal{M}' \subset\mathcal{M}$. 

Following \cite{bronstein2008not}, we model the part $\mathcal{M}'$ by means of an indicator function $v:\mathcal{M} \rightarrow \{ 0,1\}$ such that $v(x)=1$ if $x\in \mathcal{M}'$ and zero otherwise.  
Assuming that $v$ is known, the {\em partial functional correspondence} between $\mathcal{N}$ and $\mathcal{M}$ can be expressed as $Tf = vg$, where $v$ can be regarded as a kind of mask, and anything outside the region where $v=1$ should be ignored. 
Expressed w.r.t. bases $\{\phi_i\}_{i\geq 1}$ and $\{\psi_i\}_{i\geq 1}$, the partial functional correspondence takes the form $\mathbf{C} \mathbf{A} = \mathbf{B}(v)$, where 
$\mathbf{B}(v)$ denotes a matrix of weighted inner products with elements given by $b_{ij}(v) = \int_{\mathcal{M}} v(x)\psi_i(x)g_j(x) dx$ (when $v(x)\equiv 1$, $\mathbf{B}$ is simply the matrix of Fourier coefficients defined in~\eqref{eq:funcorr1}).

This brings us to the problem we are considering throughout this paper, involving optimization w.r.t. correspondence (encoded by the coefficients $\mathbf{C}$) and the part $v$, 
\begin{eqnarray}
\label{eq:funcorr_part1}
\min_{\mathbf{C}, v} \,
\| \mathbf{C}\mathbf{A} - \mathbf{B}(\eta(v)) \|_{2,1} + \rho_{\mathrm{corr}}(\mathbf{C}) + \rho_{\mathrm{part}}(v)\,,
\end{eqnarray}
where $\eta(t) = \tfrac{1}{2}\left( \tanh(2t-1) +1\right)$ saturates the part indicator function between zero and one (see below).
Here $\rho_{\mathrm{corr}}$ and $\rho_{\mathrm{part}}$ denote regularization terms for the correspondence and the part, respectively; these terms are explained below. 
We use the $L_{2,1}$ matrix norm (equal to the sum of $L_2$-norms of matrix columns) to handle possible outliers in the corresponding data, as such a norm promotes column-sparse matrices. A similar norm was adopted in \cite{DBLP:journals/tog/HuangWG14} to handle spurious maps in shape collections.

Note that in order to avoid a combinatorial optimization over binary-valued $v$, we use a continuous $v$ with values in the range $(-\infty, +\infty)$, saturated by the non-linearity $\eta$. This way, $\eta(v)$ becomes a soft membership function with values in the range $[0,1]$.


\paragraph*{Part regularization.}

Similarly to \cite{bronstein2008not,pokrass13-nmtma}, we try to find the part with area closest to that of the query and with shortest boundary. This can be expressed as 
\begin{eqnarray}
\label{eq:funcorr_reg1}
\rho_{\mathrm{part}}(v) &=& \mu_1\left( \mathrm{area}(\mathcal{N}) - \int_{\mathcal{M}} \eta(v) dx \right)^2 \\
&+& \mu_2 \int_{\mathcal{M}} \xi(v)\|\nabla_{\mathcal{M}}v \| dx\,,  \nonumber
\end{eqnarray}
where $\xi(t) \approx \delta \left(\eta(t)-\tfrac{1}{2}\right)$ and the norm is on the tangent space.  
The $\mu_2$-term in~(\ref{eq:funcorr_reg1}) is an intrinsic version of the {\em Mumford-Shah functional} \cite{mumford1989optimal}, measuring the length of the boundary of a part represented by a (soft) membership function.  
This functional was used previously in image segmentation applications \cite{vese2002multiphase}.

\begin{figure}[t]
\centering
\begin{minipage}[b]{0.49\linewidth}
\centering
 \begin{overpic}
 [trim=0cm 0cm 0cm 0cm,clip,width=1.0\linewidth]{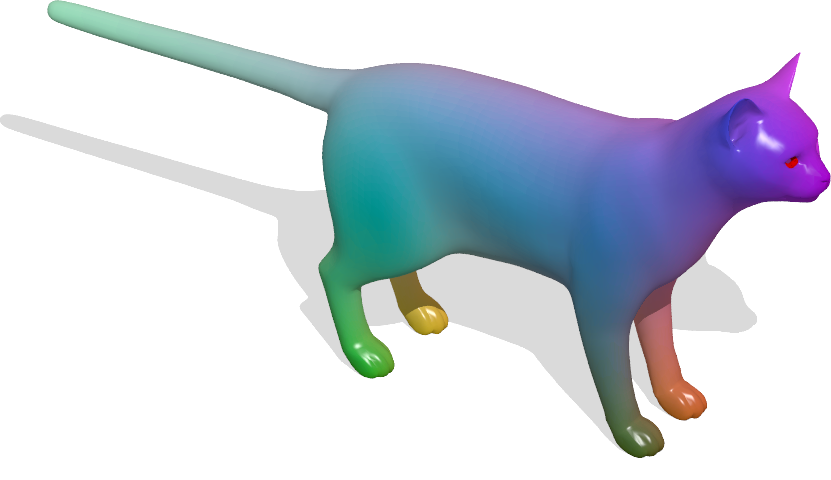}
 \put(52.0,4){\footnotesize $\mathcal{M}$}
 \end{overpic}
\end{minipage}
\begin{minipage}[b]{0.49\linewidth}
\centering
 \begin{overpic}
   [trim=0cm 0cm 0cm 0cm,clip,width=0.57\linewidth]{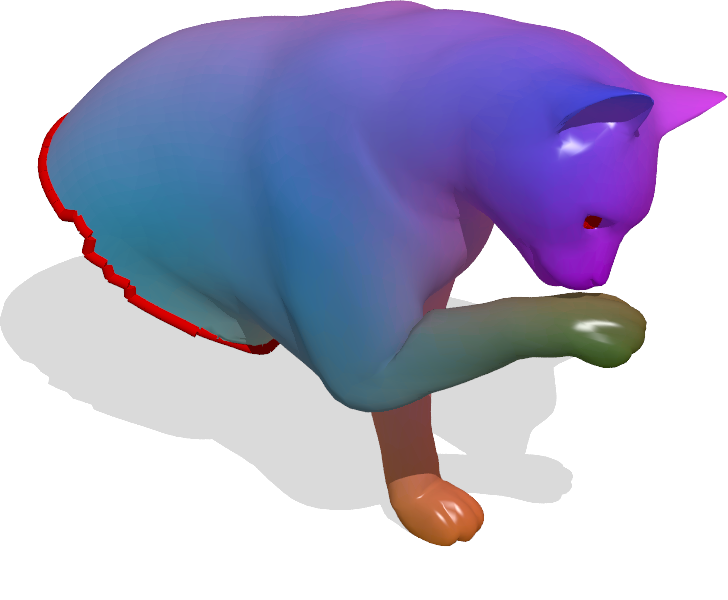}
   \put(30.0,4){\footnotesize $\mathcal{N}$}
   \end{overpic}
\end{minipage}
\begin{minipage}{0.49\linewidth}
\centering
\vspace{2pt}
\includegraphics[width=0.65\linewidth]{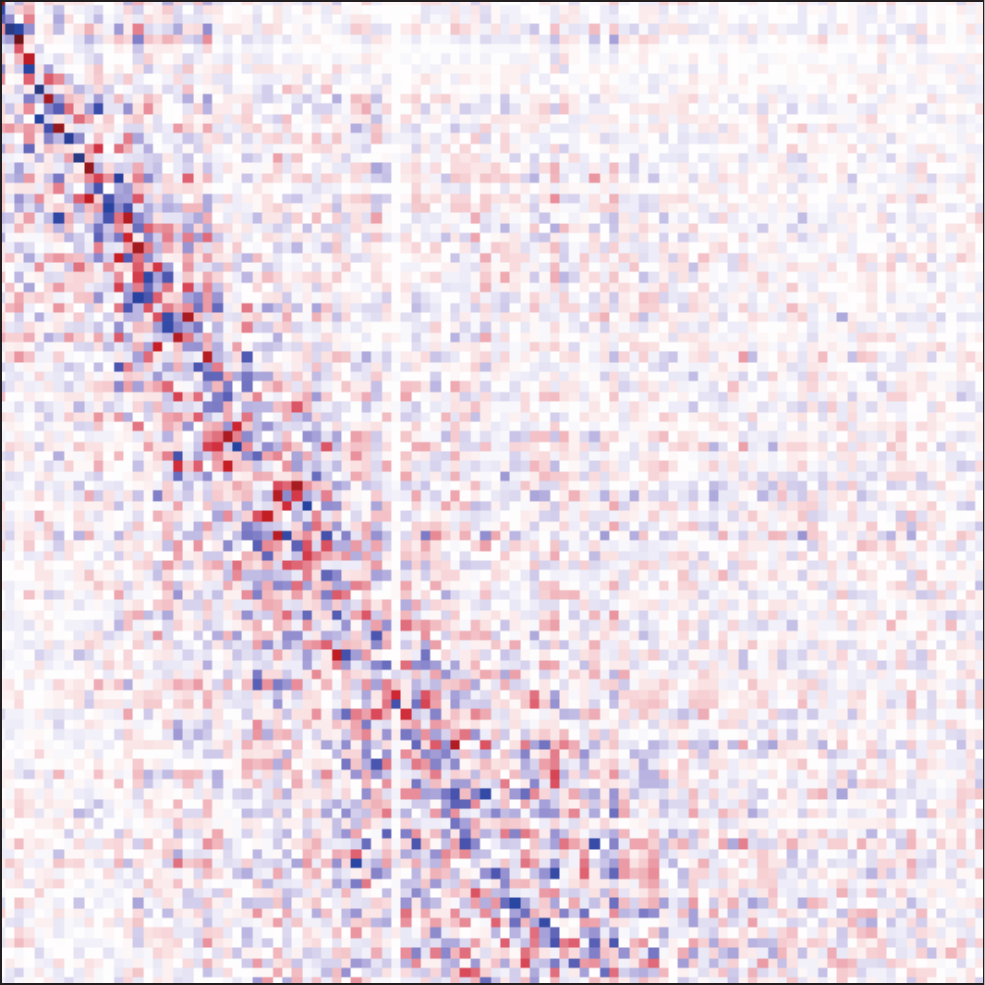}

$\mathbf{C}$
\end{minipage}\hspace{-10pt}
\begin{minipage}{0.49\linewidth}
\centering
\vspace{2pt}
\includegraphics[width=0.65\linewidth]{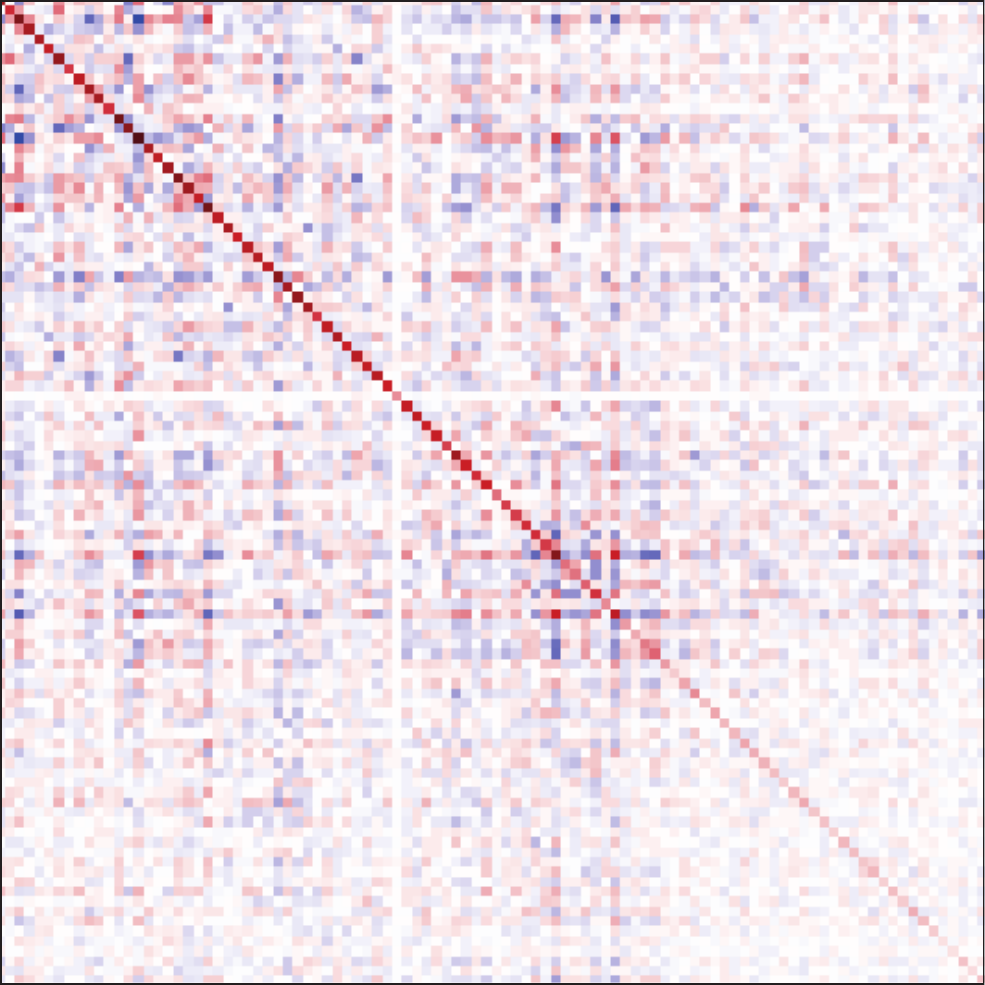}

$\mathbf{C}\T\mathbf{C}$
\end{minipage}
\begin{minipage}{0.49\linewidth}
\centering
\vspace{2pt}
\includegraphics[width=0.65\linewidth]{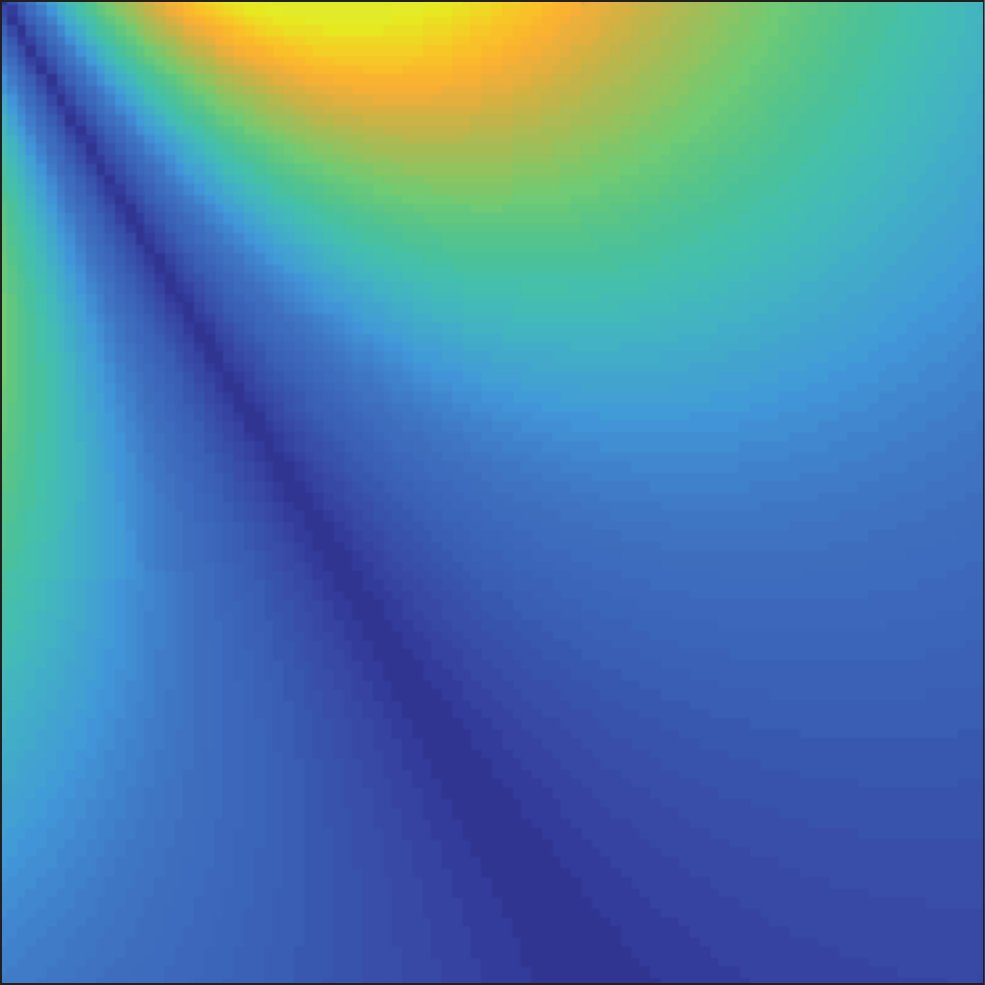}

$\mathbf{W}$
\end{minipage}\hspace{-10pt}
\begin{minipage}{0.49\linewidth}
\centering
\vspace{2pt}
%
%
\definecolor{mycolor1}{rgb}{0.00000,0.44700,0.74100}%
\begin{tikzpicture}

\begin{axis}[%
width=0.65\linewidth,
scale only axis,
separate axis lines,
every outer x axis line/.append style={black},
every x tick label/.append style={font=\color{black}, font=\tiny},
xmin=0,
xmax=100,
xlabel near ticks,
ylabel near ticks,
every outer y axis line/.append style={black},
every y tick label/.append style={font=\color{black}, font=\tiny},
ymin=0,
ymax=4.5,
ylabel={\footnotesize Singular value},
xlabel={\footnotesize $\sigma$},
axis background/.style={fill=white}
]
\addplot [color=mycolor1,solid,line width=1.8pt,forget plot]
  table[row sep=crcr]{%
1	4.25305322484127\\
2	3.86701095589921\\
3	2.63345182839754\\
4	2.56490790257978\\
5	2.47271521896802\\
6	2.38228890327154\\
7	2.33559609507753\\
8	2.29532036973249\\
9	2.21283221503835\\
10	2.18037972949379\\
11	2.14667166986774\\
12	2.13260199859436\\
13	2.05758327817008\\
14	2.01933498807363\\
15	1.92164523137034\\
16	1.88850418119439\\
17	1.74031871464691\\
18	1.6990677044636\\
19	1.68018517538559\\
20	1.65158851632567\\
21	1.64761521211956\\
22	1.60585492207701\\
23	1.57628332059022\\
24	1.53694856703386\\
25	1.4662419964211\\
26	1.41537028467486\\
27	1.39399122005319\\
28	1.36157659841947\\
29	1.3421314293627\\
30	1.31505925730542\\
31	1.28449177483357\\
32	1.26614569867799\\
33	1.22452145675716\\
34	1.20961403896437\\
35	1.17037328710588\\
36	1.15637988407727\\
37	1.13289761262201\\
38	1.10860034371595\\
39	1.09639592794616\\
40	1.07413693996837\\
41	1.05366989071716\\
42	1.03416672082685\\
43	1.02379910223642\\
44	0.983266857312993\\
45	0.968432603670871\\
46	0.94942882490028\\
47	0.945513249494663\\
48	0.939831041216294\\
49	0.92674947982629\\
50	0.92055329098213\\
51	0.906506318334109\\
52	0.902195884199385\\
53	0.892987633085877\\
54	0.880921375710316\\
55	0.869148445774887\\
56	0.862620003450565\\
57	0.854426638763496\\
58	0.742487243994358\\
59	0.705041678981068\\
60	0.572588467340212\\
61	0.470051120607331\\
62	0.352601698640107\\
63	0.294482539605524\\
64	0.241097882421209\\
65	0.187664789687436\\
66	0.157987060381702\\
67	0.123022147988541\\
68	0.0944603731819541\\
69	0.048136458569807\\
70	0.0416187087024577\\
71	0.0244594605356658\\
72	0.0214094282796877\\
73	0.0178183974429989\\
74	0.0106117447751113\\
75	0.0088028294310205\\
76	0.00650946791408145\\
77	0.00266157466653922\\
78	0.00250400854770028\\
79	0.00172588318668932\\
80	0.00166294760110711\\
81	0.000735302806401513\\
82	0.000427439260722509\\
83	0.000380643725078989\\
84	0.00019108112377644\\
85	0.000181102388917159\\
86	6.34132355308187e-005\\
87	5.8675783096837e-005\\
88	2.98137564482039e-005\\
89	2.58918004929057e-005\\
90	8.34452413595101e-006\\
91	4.01405794825589e-006\\
92	2.60145761201873e-006\\
93	2.26917704206539e-006\\
94	1.73167186864753e-007\\
95	3.34081914520057e-008\\
96	1.98925815008838e-008\\
97	6.42816047093916e-009\\
98	2.52536334761187e-010\\
99	1.41636116693071e-012\\
100	4.18287913640926e-014\\
};
\end{axis}
\end{tikzpicture}%
\end{minipage}
\caption{\label{fig:C}\rev{A partial functional map $\vct{C}$ from $\mathcal{N}$ to $\mathcal{M}$ has a slanted-diagonal structure (second row, left). The low-rank nature of such a map is also manifested in its singular values (bottom right). If the map is volume-preserving, then its full-rank sub-matrix is orthogonal: observe how $\vct{C}\T\vct{C}$ approximates the identity, with a trail of small values along the diagonal corresponding to the almost-zero block of $\vct{C}$.}}
\end{figure}

\paragraph*{Correspondence regularization.} 
For the correspondence, we use the penalty 
\begin{eqnarray}
\label{eq:funcorr_reg2}
\rho_{\mathrm{corr}}(\mathbf{C}) &=& \mu_3\|\mathbf{C} \circ \mathbf{W}\|_\mathrm{F}^2 
+  \mu_4 \sum_{i\neq j}(\mathbf{C}^\top\mathbf{C})_{ij}^2 \nonumber\\
&+&  \mu_5 \sum_{i}((\mathbf{C}^\top\mathbf{C})_{ii} - d_i)^2\,,
\end{eqnarray}
where $\circ$ denotes Hadamard (element-wise) matrix product. 
The $\mu_3$-term models the special slanted-diagonal structure of $\mathbf{C}$ that we observe in partial matching problems (see Fig.~\ref{fig:C}); 
%
%
the theoretical motivation for this behavior was presented in Sec.~\ref{sec:perturb}. Here, $\mathbf{W}$ is a weight matrix with zeros along the slanted diagonal and large values outside (see Fig.~\ref{fig:C}; details on the computation of $\mathbf{W}$ are provided in Appendix A).

The $\mu_4$-term promotes orthogonality of $\mathbf{C}$ by penalizing the off-diagonal elements of $\mathbf{C}^\top\mathbf{C}$. The reason is that for isometric shapes, the functional map is volume-preserving, and this is manifested in orthogonal $\mathbf{C}$ \cite{ovsjanikov12}. Note that differently from the classical case (\ie, full shapes), in our setting we can only require area preservation going in the direction from partial to complete model, as also expressed by the $\mu_1$-term in \eqref{eq:funcorr_reg1}. For this reason, we do not impose any restrictions on $\mathbf{C}\mathbf{C}\T$ and we say that the matrix is {\em semi}-orthogonal.

Finally, note that due to the low-rank nature of $\mathbf{C}$ we can not expect the product $\mathbf{C}\T\mathbf{C}$ to be full rank. Indeed, we expect elements off the slanted diagonal of $\mathbf{C}$ to be close to zero and thus $\mathbf{C}\T\mathbf{C}\approx\begin{pmatrix}\mathbf{I}&\mathbf{0}\\\mathbf{0}&\mathbf{0}\end{pmatrix}$. The $\mu_5$-term in \eqref{eq:funcorr_reg2} models this behavior, where vector $\mathbf{d} = (d_1, \hdots, d_k)$ determines how many singular values of $\mathbf{C}$ are non-zero (the estimation of $\mathbf{d}$ is straightforward, and described in Appendix A). 


\paragraph*{Remark.} 
The fact that matrix $\vct{C}$ is low-rank is a direct consequence of partiality. This can be understood by recalling from Eq.~\eqref{eq:funcorr1} that the (non-truncated) functional map representation amounts to an orthogonal change of basis; since in the standard basis the correspondence matrix is low-rank (as it contains zero-sum rows), this property is preserved by the change of basis.

\noindent
In Fig.~\ref{fig:C} we show an example of a ground-truth partial functional map $\vct{C}$, illustrating its main properties.

%
%

\begin{figure*}[t]
  \centering
  \begin{overpic}
  [trim=0cm 0cm 0cm 0cm,clip,width=0.71\linewidth]{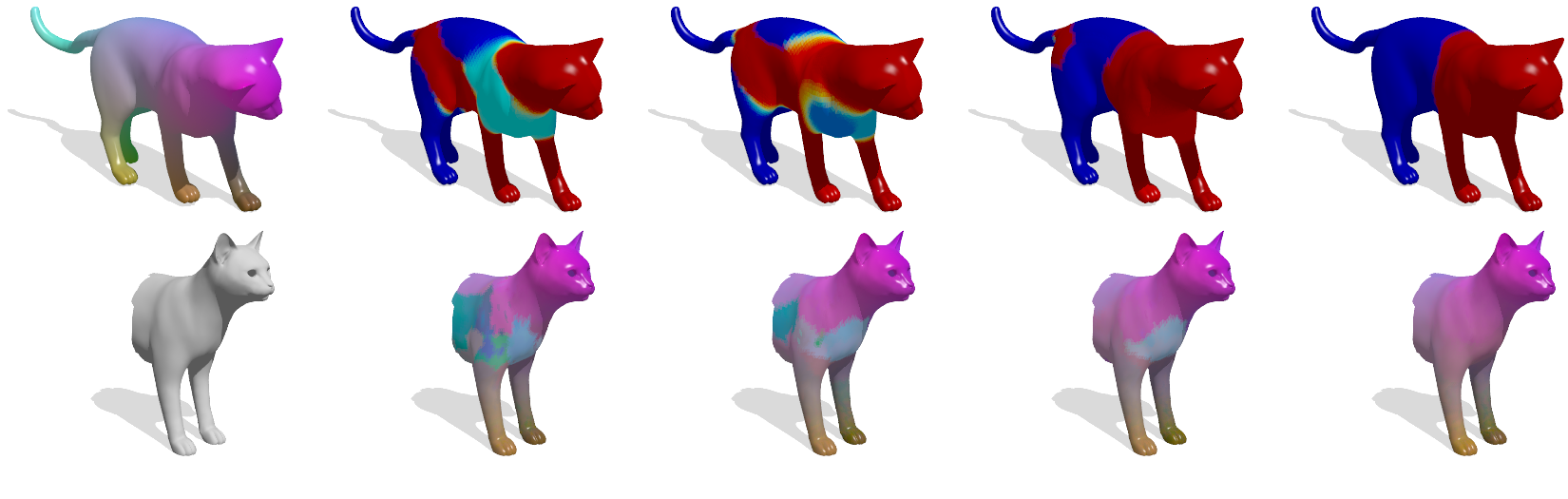}
  \put(28.5,0){\footnotesize Iteration 1}
  \put(52,0){\footnotesize 2}
  \put(73,0){\footnotesize 3}
  \put(93,0){\footnotesize 4}
  \put(19.7,3.3){\includegraphics[width=0.06\linewidth]{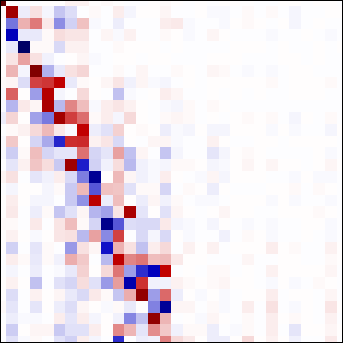}}
  \put(40.1,3.3){\includegraphics[width=0.06\linewidth]{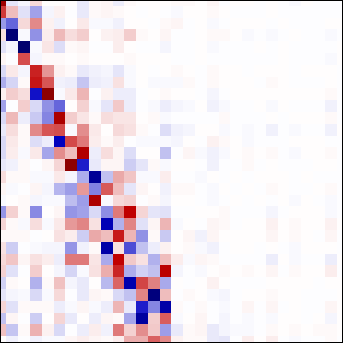}}
  \put(60.6,3.3){\includegraphics[width=0.06\linewidth]{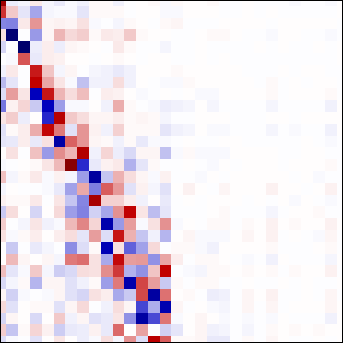}}
  \put(81.0,3.3){\includegraphics[width=0.06\linewidth]{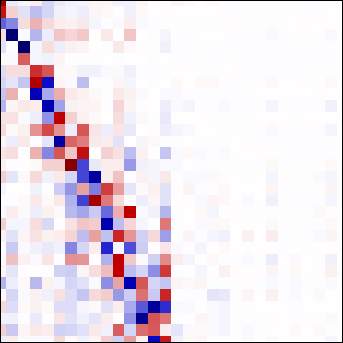}}
  \end{overpic}
  \hspace{5pt}
  \setlength\figureheight{2.3cm} 
  \setlength\figurewidth{0.4\columnwidth}    
%
%
\definecolor{mycolor1}{rgb}{0.00000,0.44700,0.74100}%
\definecolor{mycolor2}{rgb}{0.85000,0.32500,0.09800}%
\begin{tikzpicture}

\begin{axis}[%
width=0.95092\figurewidth,
height=\figureheight,
at={(0\figurewidth,0\figureheight)},
scale only axis,
unbounded coords=jump,
separate axis lines,
xlabel near ticks,
ylabel near ticks,
every outer x axis line/.append style={black},
every x tick label/.append style={font=\color{black}, font=\footnotesize},
xtick pos=left,
xmin=0,
xmax=108,
xlabel={\footnotesize Iteration},
every outer y axis line/.append style={black},
every y tick label/.append style={font=\color{black}, font=\footnotesize},
ymode=log,
ymin=10000,
ymax=10000000000,
yminorticks=true,
ylabel={\footnotesize Energy},
axis background/.style={fill=white},
legend style={at={(0.97,0.94)}, legend cell align=left,align=left,draw=white!15!black}
]
\addplot [color=mycolor1,solid,line width=2.5pt]
  table[row sep=crcr]{%
1	1220621356.55724\\
2	1164021309.69353\\
3	1065769726.32111\\
4	915910750.351788\\
5	733770445.151518\\
6	578610217.527309\\
nan	nan\\
13	186167767.208831\\
14	171954117.056176\\
15	149191509.881573\\
16	119116262.208994\\
17	90081226.2718551\\
18	71871066.9726561\\
nan	nan\\
25	12110980.4632078\\
26	10024837.7898236\\
27	7235432.51908126\\
28	4473358.65034454\\
29	2637879.81705488\\
30	1795403.00212644\\
nan	nan\\
37	1504423.68750492\\
38	1347484.63758174\\
39	1194852.74092278\\
40	1095177.09467572\\
41	1051805.88371942\\
42	1037892.22943382\\
nan	nan\\
49	294080.622907255\\
50	290248.470326373\\
51	289051.419696822\\
52	288650.558312891\\
53	288320.275145748\\
54	287998.226755654\\
nan	nan\\
61	64384.0685806945\\
62	64115.6323545185\\
63	63871.6740191523\\
64	63284.8820902961\\
65	62867.7783997044\\
66	62235.3139315868\\
nan	nan\\
73	49772.4547290202\\
74	49293.8784843479\\
75	48809.3819658365\\
76	48550.7770799538\\
77	48260.0802016675\\
78	48061.1078393638\\
nan	nan\\
85 39845.5698441863\\
86 39801.2334463533\\
87 39768.0548224052\\
88	39724.5864609821\\
89	39704.0266311963\\
90	39690.2983659519\\
nan	nan\\
97		36471.4479522407\\
98		36463.4545648728\\
99		36451.5115995924\\
100	36441.8655149333\\
101	36429.6612899553\\
102	36424.9343924512\\
};
\addlegendentry{\footnotesize{$\mathbf{C}$-step}};

\addplot [color=mycolor2,solid,line width=2.5pt]
  table[row sep=crcr]{%
7	578608231.65176\\
8	543951318.071113\\
9	481983090.415574\\
10	391832154.04445\\
11	271976297.19333\\
12	186167824.938903\\
nan	nan\\
19	71869575.6110532\\
20	61766402.8981854\\
21	44328382.3437814\\
22	24182496.4140589\\
23	13194164.4743132\\
24	12111010.5060428\\
nan	nan\\
31	1794196.96217395\\
32	1665381.56921612\\
33	1653478.98748827\\
34	1630859.66206835\\
35	1586858.53513712\\
36	1504425.12694203\\
nan	nan\\
43	1036774.0416576\\
44	954060.006601227\\
45	637961.523916208\\
46	565515.380623351\\
47	404573.140273051\\
48	294075.005036848\\
nan	nan\\
55	287310.032818963\\
56	248073.562969686\\
57	199099.851456765\\
58	127602.508654988\\
59	86667.2801113346\\
60	64383.339430432\\
nan	nan\\
67	62296.2426998103\\
68	57477.0352211415\\
69	55619.1849068175\\
70	53714.4579493934\\
71	51822.4243819838\\
72	49772.3941863437\\
nan	nan\\
79	48116.5508177516\\
80	47286.8979513248\\
81	46739.5144472637\\
82	45828.9790369717\\
83	44087.973419497\\
84	39845.5383317312\\
nan	nan\\
91	39677.5579158372\\
92	37793.1954112274\\
93	37343.676801516\\
94	36912.4360919744\\
95	36666.6279513473\\
96	36471.4427323419\\
nan	nan\\
103	36426.2472244287\\
104	36338.116452901\\
105	36252.6028194372\\
106	36115.8278851961\\
107	35887.7733442575\\
108	35464.3291099624\\
};
\addlegendentry{\footnotesize{$\nu$-step}};

\end{axis}

\begin{axis}[%
width=0.95092\figurewidth,
height=\figureheight,
at={(0\figurewidth,0\figureheight)},
scale only axis,
every outer x axis line/.append style={black},
every x tick label/.append style={font=\color{black}, font=\footnotesize},
xmin=0,
xmax=27.0,
xlabel={\footnotesize Time (sec.)},
xlabel near ticks,
xtick={0,5,10,15,20,25,30},
xticklabels={0,5,10,15,20,25,30},
every outer y axis line/.append style={black},
every y tick label/.append style={font=\color{black}},
ymin=0,
ymax=1,
ytick={\empty},
axis x line*=top,
axis y line=none
]
\end{axis}

\end{tikzpicture}%
  \caption{\label{fig:alternating}An example of the matching process operating on two shapes from TOSCA. The algorithm alternatingly optimizes over corresponding part (top row) and functional correspondence (bottom row). Corresponding points between full and partial shape are shown with the same color. This solution was obtained by using 30 eigenfunctions on both manifolds.}
\end{figure*}

\paragraph*{Alternating scheme.}
To solve the optimization problem~(\ref{eq:funcorr_part1}), we perform an alternating optimization w.r.t. to $\mathbf{C}$ and $v$, repeating the following steps until convergence: 

{\em C-step: } Fix $v^\ast$, solve for correspondence $\mathbf{C}$
\begin{eqnarray}
\label{eq:funcorr_part_C}
\min_{\mathbf{C}} \,
\| \mathbf{C}\mathbf{A} - \mathbf{B}(\eta(v^\ast)) \|_{2,1} + \rho_{\mathrm{corr}}(\mathbf{C})\,.
\end{eqnarray}

{\em V-step: } Fix $\mathbf{C}^\ast$, solve for part $v$
\begin{eqnarray}
\label{eq:funcorr_part_v}
\min_{v} \,
\| \mathbf{C}^\ast\mathbf{A} - \mathbf{B}(\eta(v)) \|_{2,1}  + \rho_{\mathrm{part}}(v)\,.
\end{eqnarray}
A practical example of the alternating scheme applied to a pair of shapes is shown in Fig.~\ref{fig:alternating}.

%
%

\section{Implementation} 
\label{sec:impl}

We refer to Appendix A for the discretization of the regularization terms appearing in \eqref{eq:funcorr_reg1} and \eqref{eq:funcorr_reg2}.

%
%

\paragraph*{Numerical optimization.}
We implemented our matching framework in Matlab/C++ using the manifold optimization toolbox~\cite{manopt}. Each optimization step was performed by the method of nonlinear conjugate gradients. Detailed derivations of the involved gradients can be found in Appendix B. We initialize the alternating scheme by fixing $\vct{v}^\ast = \vct{1}$ (a vector of $m$ ones), ${\vct{C} = \vct{W}}$, and by optimizing over $\vct{C}$. In all our experiments we observed convergence in 3-5 outer iterations (around 5 mins. for a pair of shapes).


\paragraph*{Refinement.}

In order to account for noisy data, we also run a refinement step after each {\em C-step}. Specifically, assume $\vct{C}^\ast$ is a local optimum of problem \eqref{eq:funcorr_part_C}, \rev{and consider the term $\| \vct{C}^\ast \bm{\Phi}\T - \bm{\Psi}\T \bm{\Pi} \|_{F}$, where $\bm{\Pi}$ is a left-stochastic binary matrix assigning each column of $\bm{\Psi}\T$ to the nearest column of $\vct{C}^\ast \bm{\Phi}\T$; this is done by $n$ nearest-neighbor searches in $\mathbb{R}^k$, one per column. Given the optimal $\bm{\Pi}^\ast$, we solve for the map $\vct{C}$ minimizing $\| \vct{C} \bm{\Phi}\T - \bm{\Psi}\T \bm{\Pi}^\ast \|_{F}$ plus the $\mu_4$, $\mu_5$ terms of Eq.~\eqref{eq:funcorr_reg2}. We alternate the $\vct{C}^\ast$ and $\bm{\Pi}^\ast$ steps until convergence. This refinement step can be seen as a generalization to partial maps of the ICP-like technique found in~\cite{ovsjanikov12}, and can be interpreted as an attempt to improve the alignment between the spectral embeddings of the two shapes. Further note that matrix $\bm{\Pi}^\ast$ encodes the point-wise correspondence between $\mathcal{N}$ and $\mathcal{M}$, which is used to evaluate the accuracy of our method.}

%




\section{Experimental results}
\label{sec:exp}
%

%
%
%

\paragraph*{Datasets.}
%
As base models, we use shapes from the TOSCA dataset~\cite{bbk08}, consisting of 76 nearly-isometric shapes subdivided into 8 classes.
 Each class comes with a ``null'' shape in a standard pose (extrinsically bilaterally symmetric), and ground-truth correspondences are provided for all shapes within the same class. 
In order to make the datasets more challenging and avoid compatible triangulations, all shapes were remeshed to 10K vertices by iterative pair contractions~\cite{garland97}. Then, missing parts were introduced in the following ways:\footnote{The datasets together with code for our method are available for download at {\small \url{http://vision.in.tum.de/data/datasets/partial}}.
}

{\em Regular cuts}. The null shape of each class was cut with a plane at 6 different orientations, including an exact cut along the symmetry plane. The six cuts were then transferred to the remaining poses using the ground-truth correspondence, resulting in 456 partial shapes in total. Some examples are shown in Fig.~\ref{fig:spectra} and \ref{fig:C}.

\begin{figure}[t]
  \centering
  \centering
  \setlength\figureheight{3cm} 
\setlength\figurewidth{0.85\columnwidth}  
%
%
\definecolor{mycolor0}{rgb}{0.00000,0.00000,0.00000}%
\definecolor{mycolor1}{rgb}{0.00000,0.44700,0.74100}%
\definecolor{mycolor2}{rgb}{0.85000,0.32500,0.09800}%
\definecolor{mycolor3}{rgb}{0.92900,0.69400,0.12500}%
\begin{tikzpicture}

\begin{axis}[%
width=0.95092\figurewidth,
height=\figureheight,
scale only axis,
separate axis lines,
every outer x axis line/.append style={black},
every x tick label/.append style={font=\color{black}, font=\footnotesize},
xlabel near ticks,
ylabel near ticks,
xmin=20,
xmax=80,
xlabel={\footnotesize Partiality (\%)},
every outer y axis line/.append style={black},
every y tick label/.append style={font=\color{black}, font=\footnotesize},
ymin=0.05,
ymax=0.6,
ymajorgrids,
ylabel={\footnotesize Mean geodesic error},
axis background/.style={fill=white},
legend style={at={(0.02,0.4)},anchor=south west,legend cell align=left,align=left,draw=white!15!black}
]
\addplot [color=mycolor0,solid,line width=1.5pt]
  table[row sep=crcr]{%
20	0.0673317570531928\\
35	0.0826383736907717\\
50	0.0943630878200045\\
65	0.0810068407735256\\
80	0.0939021257181005\\
};
\addlegendentry{\footnotesize Ours};

\addplot [color=mycolor1,solid,line width=1pt]
  table[row sep=crcr]{%
20	0.210486405527758\\
35	0.24561942468432\\
50	0.265323198742567\\
65	0.365648913335047\\
80	0.357423452803931\\
};
\addlegendentry{\footnotesize Func. maps};

\addplot [color=mycolor2,solid,line width=1pt]
  table[row sep=crcr]{%
20	0.27328131758331\\
35	0.31582532525776\\
50	0.395736537602761\\
65	0.42857385628857\\
80	0.43365282875285\\
};
\addlegendentry{\footnotesize Elastic net};

\addplot [color=mycolor3,solid,line width=1pt]
  table[row sep=crcr]{%
20	0.354395740758101\\
35	0.39234495739579\\
50	0.465865295683825\\
65	0.555673856826597\\
80	0.59546148641864\\
};
\addlegendentry{\footnotesize Voting};

\end{axis}
\end{tikzpicture}%
  \begin{overpic}
  [trim=0cm 0cm 0cm 0cm,clip,width=\linewidth]{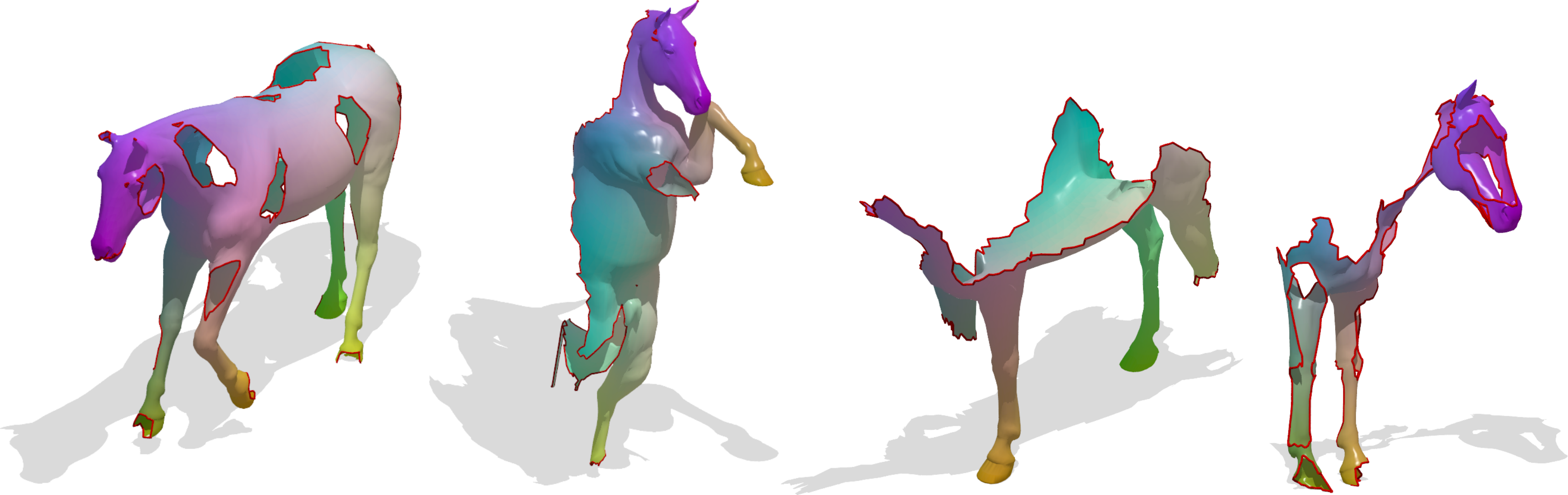}
  \end{overpic}
  \caption{\label{fig:horse_partiality} Correspondence quality (in terms of mean geodesic error, in \% of diameter) obtained by different methods at increasing levels of partiality. 
 Other methods show significant performance drop with increasing partiality, while the performance of our method is nearly-constant.  
 }
\end{figure}

{\em Irregular holes}. Given a shape and an ``area budget'' determining the fraction of area to keep (40\%, 70\%, and 90\%), we produced additional shapes by an erosion process applied to the surface.  Specifically, seed holes were placed at 5, 25, and 50 farthest samples over the shape; the holes were then enlarged to meet the specified area budget. 
The total number of shapes produced this way was 684. Examples of this dataset are shown in Fig.~\ref{fig:horse_partiality} and \ref{fig:examples}.

{\em Range images}. We simulated range images by taking orthographic projections of the original TOSCA shapes from different viewpoints. Each range image was produced via ray casting from a regular grid with a resolution of ${100\times150}$ pixels. Examples are shown in Fig. \ref{fig:examples}.

{\em Point clouds}. Point clouds were generated by taking a subset of shapes from the first two datasets. Each partial shape was then resampled uniformly to 1000 farthest points, and the tessellation removed. See Fig. \ref{fig:examples} for examples.

Where not specified otherwise, we use 120 random partial shapes for the first dataset and 80 for the second, equally distributed among the different classes. Each partial shape is then matched to the null shape of the corresponding class.


%
%

\paragraph*{Error measure.}
For the evaluation of the correspondence quality, we used the 
Princeton benchmark protocol \cite{DBLP:journals/tog/KimLF11} for point-wise maps. Assume that a correspondence algorithm produces a pair of points $(x,y) \in \mathcal{N} \times \mathcal{M}$, whereas the ground-truth correspondence is $(x,y^*)$. Then, the inaccuracy of the correspondence is measured as  
\begin{eqnarray}
\epsilon(x) &=& \frac{d_\mathcal{M}(y,y^*)}{  \mathrm{area}(\mathcal{M})^{1/2} }, 
\label{eq:harderr}
\end{eqnarray}
and has units of normalized length on $\mathcal{M}$ (ideally, zero). Here $d_{\mathcal{M}}$ is the geodesic distance on $\mathcal{M}$.  
The value $\epsilon(x)$ is averaged over all shapes $\mathcal{N}$. 
We plot cumulative curves showing the percent of matches which have error smaller than a variable threshold. 

\paragraph*{Methods.}
We compared the proposed method with (full) functional maps \cite{ovsjanikov12}, elastic net \cite{rodola13iccv}, and the voting method of \cite{sahilliouglu2014partial} using the code provided by the respective authors. 

\begin{figure}[t]
  \centering
\setlength\figureheight{4cm} 
\setlength\figurewidth{0.85\columnwidth}  
%
%
\definecolor{mycolor0}{rgb}{0.00000,0.00000,0.00000}%
\definecolor{mycolor1}{rgb}{0.00000,0.44700,0.74100}%
\definecolor{mycolor2}{rgb}{0.85000,0.32500,0.09800}%
\definecolor{mycolor3}{rgb}{0.92900,0.69400,0.12500}%
\begin{tikzpicture}

\begin{axis}[%
width=0.95092\figurewidth,
height=\figureheight,
at={(0\figurewidth,0\figureheight)},
scale only axis,
separate axis lines,
every outer x axis line/.append style={black},
every x tick label/.append style={font=\color{black}, font=\footnotesize},
xmin=0,
xmax=0.25,
xlabel={\footnotesize{Geodesic error}},
xlabel near ticks,
xtick={0,0.05,0.1,0.15,0.2,0.25},
xticklabels={0,0.05,0.1,0.15,0.2,0.25},
ylabel near ticks,
every outer y axis line/.append style={black},
every y tick label/.append style={font=\color{black}, font=\footnotesize},
ymin=0,
ymax=100,
xmajorgrids,
ymajorgrids,
ylabel={\footnotesize{\% Correspondences}},
axis background/.style={fill=white},
legend style={at={(0.975,0.05)},anchor=south east,legend cell align=left,align=left,draw=black}
]
\addplot [color=mycolor0,solid,line width=1.5pt]
  table[row sep=crcr]{%
0	24.2572163281626\\
0.01	36.8674805962844\\
0.02	55.4124150662279\\
0.03	66.5013849262995\\
0.04	73.2913684050946\\
0.05	77.8697831073199\\
0.06	81.1091715595663\\
0.07	83.4222371441168\\
0.08	85.0368234601316\\
0.09	86.2361226089073\\
0.1	87.1398480977064\\
0.11	87.8343423071709\\
0.12	88.3984859640127\\
0.13	88.88833201268\\
0.14	89.3003519088588\\
0.15	89.6687015527815\\
0.16	89.9994774835807\\
0.17	90.3108621992272\\
0.18	90.6094737822812\\
0.19	90.8896154524348\\
0.2	91.1603317863416\\
0.21	91.4412033207274\\
0.22	91.7332478215196\\
0.23	92.007244335827\\
0.24	92.2653241913704\\
0.25	92.4793625599221\\
};
\addlegendentry{\footnotesize Ours};

\addplot [color=mycolor1,solid,line width=1pt]
  table[row sep=crcr]{%
0	13.2269392016424\\
0.01	20.2629822246199\\
0.02	31.4411185704861\\
0.03	39.4596140955424\\
0.04	44.7960637165293\\
0.05	48.3849234720474\\
0.06	50.8297187086699\\
0.07	52.5078945374023\\
0.08	53.6239149495817\\
0.09	54.3746541199834\\
0.1	54.9358025931746\\
0.11	55.3563307440707\\
0.12	55.7208241647083\\
0.13	56.0673102834961\\
0.14	56.3856924108301\\
0.15	56.7063854142448\\
0.16	57.0164921673832\\
0.17	57.3609005366257\\
0.18	57.7460645607963\\
0.19	58.1887877415497\\
0.2	58.6658862772535\\
0.21	59.2160780199392\\
0.22	59.8096722589892\\
0.23	60.4025536004246\\
0.24	60.9678482156638\\
0.25	61.545528456326\\
};
\addlegendentry{\footnotesize Func. maps};

\addplot [color=mycolor2,solid,line width=1pt]
  table[row sep=crcr]{%
0	3.8577\\
0.01	4.8431\\
0.02	5.4922\\
0.03	7.7555\\
0.04	11.2712\\
0.05	14.806\\
0.06	18.1318\\
0.07	23.3769\\
0.08	28.1462\\
0.09	33.1971\\
0.1	35.9235\\
0.11	36.7948\\
0.12	38.4171\\
0.13	39.0502\\
0.14	39.1344\\
0.15	39.7387\\
0.16	40.7816\\
0.17	41.8655\\
0.18	43.8952\\
0.19	44.2869\\
0.2	46.5898\\
0.21	47.5456\\
0.22	49.7463\\
0.23	50.8094\\
0.24	51.8547\\
0.25	53.376\\
};
\addlegendentry{\footnotesize Elastic net};

\addplot [color=mycolor3,solid,line width=1pt]
  table[row sep=crcr]{%
0	0.35232902109775\\
0.01	1.85193395388503\\
0.02	2.42580777419962\\
0.03	3.33395631824837\\
0.04	5.32052477727787\\
0.05	6.31315323521858\\
0.06	7.70380658348513\\
0.07	9.50991022279377\\
0.08	10.9771675871882\\
0.09	12.6739430816509\\
0.1	15.2520182232585\\
0.11	17.0856180917407\\
0.12	18.8520379623051\\
0.13	20.3548874656155\\
0.14	22.564694891571\\
0.15	24.8403285457156\\
0.16	27.4826295080083\\
0.17	28.780101700326\\
0.18	31.0091189915746\\
0.19	33.2359143430631\\
0.2	35.925156618018\\
0.21	38.2005332468367\\
0.22	40.4540421653285\\
0.23	42.5029179711372\\
0.24	44.8077047247764\\
0.25	46.9990480737554\\
};
\addlegendentry{\footnotesize Voting};

\addplot [color=mycolor0,dashed,line width=1.5pt]
  table[row sep=crcr]{%
0	38.6335290030819\\
0.01	52.8653283644154\\
0.02	70.2886026193581\\
0.03	79.4263638411683\\
0.04	84.3909725565357\\
0.05	87.4039317323874\\
0.06	89.3162680690055\\
0.07	90.5979416288766\\
0.08	91.5082095728915\\
0.09	92.2119035992635\\
0.1	92.6864517807266\\
0.11	93.052684848758\\
0.12	93.3623445003395\\
0.13	93.6196081079908\\
0.14	93.8374248995703\\
0.15	94.0383038060509\\
0.16	94.2036241666289\\
0.17	94.3594594662486\\
0.18	94.5025004825927\\
0.19	94.667714039655\\
0.2	94.8246654427973\\
0.21	94.9934599896511\\
0.22	95.1501636643483\\
0.23	95.3159220941346\\
0.24	95.4886329292746\\
0.25	95.6637006418951\\
};

\addplot [color=mycolor1,dashed,line width=1pt]
  table[row sep=crcr]{%
0	13.6937223230484\\
0.01	19.1426027509998\\
0.02	26.4890599333835\\
0.03	31.3497898183498\\
0.04	34.6626722194697\\
0.05	36.9285530831865\\
0.06	38.5411833070682\\
0.07	39.6800881325944\\
0.08	40.4929381290337\\
0.09	41.1087156971162\\
0.1	41.5662132915644\\
0.11	41.9507750816775\\
0.12	42.3263784560463\\
0.13	42.6734967489736\\
0.14	42.9777914123992\\
0.15	43.2851170458721\\
0.16	43.6328539969406\\
0.17	43.9862480208132\\
0.18	44.3972764147252\\
0.19	44.8416114876664\\
0.2	45.3405023290569\\
0.21	45.919004524926\\
0.22	46.49016333632\\
0.23	47.077346650918\\
0.24	47.6708112644055\\
0.25	48.2616609352762\\
};

\addplot [color=mycolor2,dashed,line width=1pt]
  table[row sep=crcr]{%
0	6.14970731762303\\
0.01	7.13647199197506\\
0.02	8.44577932046512\\
0.03	9.72823931486145\\
0.04	12.6753469851293\\
0.05	13.286195178339\\
0.06	14.7104307268939\\
0.07	16.5245308802627\\
0.08	20.4910697645628\\
0.09	23.7112342194339\\
0.1	26.0745507440324\\
0.11	27.3760676344513\\
0.12	29.0813527305823\\
0.13	30.0088932048243\\
0.14	31.3580561011667\\
0.15	32.8002858370299\\
0.16	34.9497424623264\\
0.17	36.1176322670464\\
0.18	38.096477625981\\
0.19	39.1400165541918\\
0.2	42.0690587672915\\
0.21	43.4058922122178\\
0.22	45.4508981862411\\
0.23	47.5951605475685\\
0.24	49.2071451738812\\
0.25	50.7376803797761\\
};

\addplot [color=mycolor3,dashed,line width=1pt]
  table[row sep=crcr]{%
0	0.0146096112763727\\
0.01	0.0518801074673678\\
0.02	0.256560993079403\\
0.03	0.522753562954579\\
0.04	0.940272708956736\\
0.05	1.42341439825732\\
0.06	2.0621152975713\\
0.07	2.80599807720565\\
0.08	3.6807130620198\\
0.09	4.44196978904461\\
0.1	5.30996763248303\\
0.11	6.22948355151086\\
0.12	7.47682936722661\\
0.13	8.59510792266838\\
0.14	9.72991497819864\\
0.15	10.9871699744935\\
0.16	12.2614354678104\\
0.17	13.4646982626744\\
0.18	14.7660951899168\\
0.19	15.9669243136004\\
0.2	17.2760411430615\\
0.21	18.5688108611781\\
0.22	20.0062962021922\\
0.23	21.3519942647063\\
0.24	22.66069583249\\
0.25	24.2027272681476\\
};

\end{axis}
\end{tikzpicture}%
  \caption{\label{fig:comparisons}
  Correspondence quality of different methods evaluated using the Princeton protocol on partial TOSCA shapes with regular cuts (solid) and irregular holes (dotted).}
\end{figure}
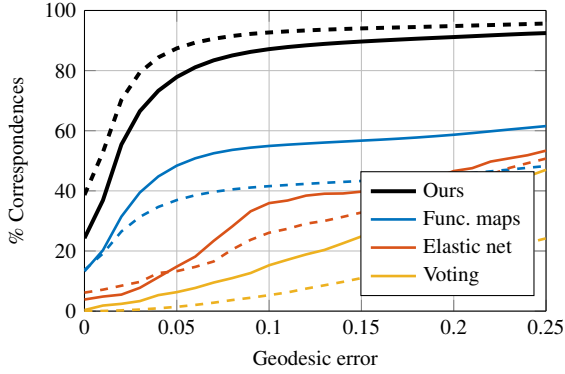
%

\paragraph*{Local descriptors. } 
\label{sec:data}
Due to the particular nature of the problem, in all our experiments we only make use of dense, {\em local} descriptors as a data term. This is in contrast with the more common scenario in which full shapes are being matched -- thus allowing to employ more robust, globally-aware features such as landmark matches, repeatable surface regions, and various spectral quantities~\cite{ovsjanikov12}.
In our experiments, we used the extrinsic SHOT~\cite{tombari10} descriptor, computed using 10 normal bins (352 dimensions in total). 
As opposed to~\cite{albarelli-pr15,pokrass13-nmtma} which ignore points close to the boundary in order to avoid boundary effects, in our formulation we retain all shape points.

%


\subsection{Sensitivity analysis}
We conducted a set of experiments aimed at evaluating the sensitivity of our approach to different parametrizations. In order to reduce overfitting we only used a subset of TOSCA (regular cuts), namely composed of the {\em cat} and {\em victoria} shape classes (20 pairs).

\begin{figure}[b]
  \centering
%
%
\definecolor{mycolor0}{rgb}{0.00000,0.00000,0.00000}%
\definecolor{mycolor1}{rgb}{0.00000,0.44700,0.74100}%
\definecolor{mycolor2}{rgb}{0.85000,0.32500,0.09800}%
\definecolor{mycolor3}{rgb}{0.92900,0.69400,0.12500}%
\begin{tikzpicture}

\begin{axis}[%
width=0.8\linewidth,
height=0.5\linewidth,
scale only axis,
separate axis lines,
every outer x axis line/.append style={black},
every x tick label/.append style={font=\color{black}, font=\footnotesize},
xmin=0,
xmax=0.25,
xlabel near ticks,
ylabel near ticks,
xtick={0,0.05,0.1,0.15,0.2,0.25},
xticklabels={0,0.05,0.1,0.15,0.2,0.25},
xlabel={\footnotesize Geodesic error},
xmajorgrids,
every outer y axis line/.append style={black},
every y tick label/.append style={font=\color{black}, font=\footnotesize},
ymin=0,
ymax=100,
ylabel={\footnotesize  \% Correspondences},
ymajorgrids,
axis background/.style={fill=white},
legend style={at={(0.97,0.50)},anchor=south east,legend cell align=left,align=left,draw=white!15!black}
]
\addplot [color=mycolor0,solid,line width=1.5pt]
  table[row sep=crcr]{%
0	30.1079714079281\\
0.01	48.9287598509017\\
0.02	63.1444702316364\\
0.03	69.0391664435541\\
0.04	71.9120841701877\\
0.05	73.6882321651061\\
0.06	75.1455246250012\\
0.07	76.3359335188125\\
0.08	77.2590863423596\\
0.09	77.9271520848328\\
0.1	78.515993029071\\
0.11	79.0675921837568\\
0.12	79.4604554586294\\
0.13	79.7448769677402\\
0.14	79.9446599693927\\
0.15	80.1344387023389\\
0.16	80.3725583302977\\
0.17	80.5838242853638\\
0.18	80.8516234211175\\
0.19	81.1866138055684\\
0.2	81.5435035837381\\
0.21	81.8384413785582\\
0.22	82.1497075861662\\
0.23	82.4870196227278\\
0.24	82.8411364384637\\
0.25	83.1872859920308\\
};
\addlegendentry{Ours};

\addplot [color=mycolor0,solid,line width=1.5pt,forget plot]
  table[row sep=crcr]{%
0	23.7880324943298\\
0.01	40.2331384992453\\
0.02	57.4244686331663\\
0.03	66.4319978775318\\
0.04	71.689649874425\\
0.05	75.1567251655072\\
0.06	77.8001542110185\\
0.07	79.7084064485423\\
0.08	81.1928288518383\\
0.09	82.2688730513422\\
0.1	82.9853409109153\\
0.11	83.4582971044488\\
0.12	83.84209784111\\
0.13	84.1293427406423\\
0.14	84.3470843485862\\
0.15	84.5371417202168\\
0.16	84.7170054369681\\
0.17	84.9084151323134\\
0.18	85.0943309728607\\
0.19	85.2890128853893\\
0.2	85.4798842990247\\
0.21	85.6611584433614\\
0.22	85.8457899143174\\
0.23	86.0448501319711\\
0.24	86.2388175872573\\
0.25	86.4309525527619\\
};

\addplot [color=mycolor0,solid,line width=1.5pt,forget plot]
  table[row sep=crcr]{%
0	27.3144157935689\\
0.01	46.6036708226614\\
0.02	64.9953084895689\\
0.03	73.6447390955994\\
0.04	78.5341438956281\\
0.05	81.3971270010473\\
0.06	83.3494361001475\\
0.07	84.62922126295\\
0.08	85.6776670651658\\
0.09	86.4223613494523\\
0.1	86.9693134792571\\
0.11	87.366496918341\\
0.12	87.7038140375798\\
0.13	87.9778010975321\\
0.14	88.2119390728784\\
0.15	88.3782428896209\\
0.16	88.4893561407191\\
0.17	88.6342871369895\\
0.18	88.7585488249114\\
0.19	88.8693820783184\\
0.2	88.9717988980804\\
0.21	89.1066640987218\\
0.22	89.2415546466221\\
0.23	89.3929587302371\\
0.24	89.5581667266887\\
0.25	89.6979357928735\\
};
\addplot [color=mycolor1,solid,line width=1pt]
  table[row sep=crcr]{%
0	2.93232353293244\\
0.01	6.91619501205223\\
0.02	10.8934212336641\\
0.03	12.9868401150599\\
0.04	14.3505730985805\\
0.05	15.0871958539285\\
0.06	15.7549786439506\\
0.07	16.378349501322\\
0.08	17.0354064865821\\
0.09	17.5294676554388\\
0.1	17.9819286368841\\
0.11	18.3925549930225\\
0.12	18.7811483224118\\
0.13	19.1475390413918\\
0.14	19.5815832872274\\
0.15	20.0004929597279\\
0.16	20.5225562684109\\
0.17	21.0381314429582\\
0.18	21.4683773829542\\
0.19	21.8899892580561\\
0.2	22.3616839116258\\
0.21	22.7840159762231\\
0.22	23.1191966453375\\
0.23	23.4136273574747\\
0.24	23.729988133714\\
0.25	24.1471818416661\\
};
\addplot [color=mycolor1,solid,line width=1pt,forget plot]
  table[row sep=crcr]{%
0	3.64207156748719\\
0.01	8.82860093050341\\
0.02	14.4215873679389\\
0.03	17.6346359954442\\
0.04	19.6961603326582\\
0.05	21.0359837755827\\
0.06	22.1006434916104\\
0.07	22.9564219717081\\
0.08	23.6205722807348\\
0.09	24.1273779863196\\
0.1	24.5149324703712\\
0.11	24.8297319814845\\
0.12	25.160548113204\\
0.13	25.4474431327959\\
0.14	25.6821015943952\\
0.15	25.9819287307948\\
0.16	26.3083190988982\\
0.17	26.7262565554766\\
0.18	27.2500624495582\\
0.19	27.7678714540939\\
0.2	28.2663476767791\\
0.21	28.7589614990543\\
0.22	29.1896297317343\\
0.23	29.5960842315458\\
0.24	30.0071078222545\\
0.25	30.4204387458257\\
};
\addplot [color=mycolor1,solid,line width=1pt,forget plot]
  table[row sep=crcr]{%
0	3.44464271142257\\
0.01	8.51435113522349\\
0.02	15.0361128520286\\
0.03	18.9872673231755\\
0.04	21.6056304397067\\
0.05	23.3564099177649\\
0.06	24.7569782063494\\
0.07	25.7878008583281\\
0.08	26.5993877654582\\
0.09	27.2756401544452\\
0.1	27.8912544711152\\
0.11	28.3038886355068\\
0.12	28.6465178276155\\
0.13	29.0061819837502\\
0.14	29.3117528739078\\
0.15	29.5837652363134\\
0.16	29.9004204175874\\
0.17	30.2907357273608\\
0.18	30.5936626566725\\
0.19	30.8926671016375\\
0.2	31.2600900674674\\
0.21	31.7310134556593\\
0.22	32.2218028932091\\
0.23	32.689117179067\\
0.24	33.120926674461\\
0.25	33.5309414084768\\
};
\addlegendentry{Func. maps};

\node[rectangle,fill=white] (n1) at (100,31) {\footnotesize 50};
\node[rectangle,fill=white] (n1) at (125,25) {\footnotesize 100};
\node[rectangle,fill=white] (n1) at (150,20) {\footnotesize 150};

\node[rectangle,fill=white] (n1) at (100,76) {\footnotesize 50};
\node[rectangle,fill=white] (n1) at (125,85) {\footnotesize 100};
\node[rectangle,fill=white] (n1) at (150,91) {\footnotesize 150};

\end{axis}
\end{tikzpicture}%
  \caption{\label{fig:rank} Correspondence quality obtained on a subset of TOSCA at increasing rank (reported as labels on top of the curves). Note the opposite behavior of the baseline approach and our regularized partial matching.}
\end{figure}
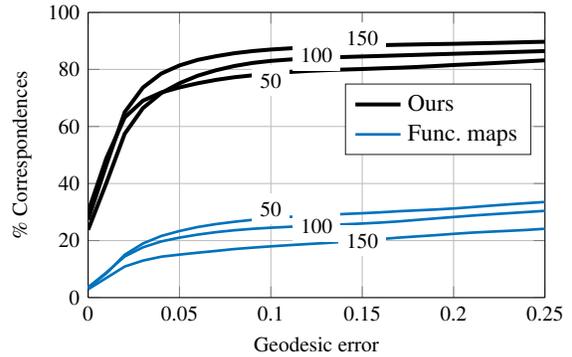
\begin{figure*}[t]
  \centering
  \begin{overpic}
  [trim=0cm 0cm 0cm 0cm,clip,width=1\linewidth]{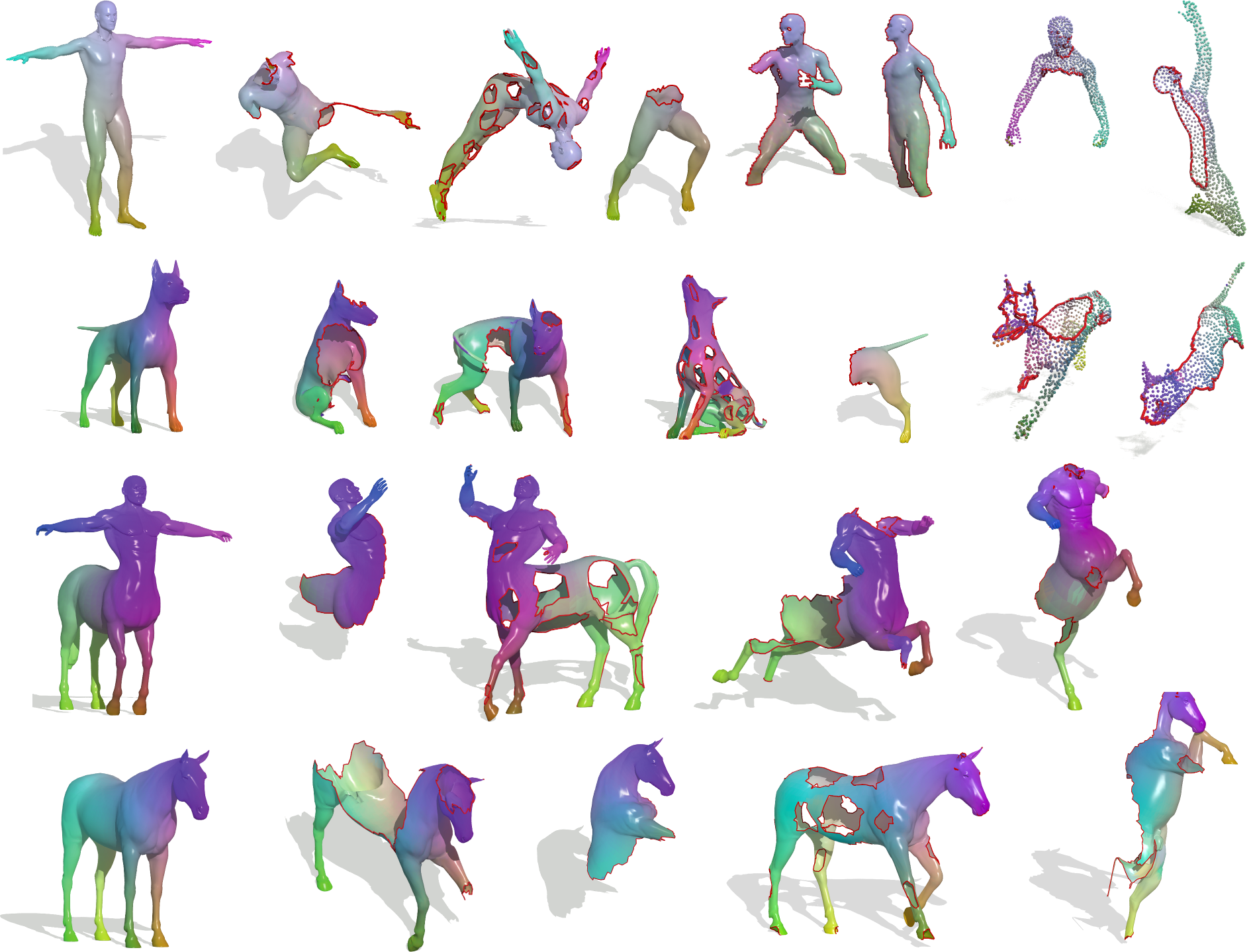}
  \put(64,58){\footnotesize range maps}
  \end{overpic}
  \caption{\label{fig:examples}Examples of partial functional correspondence obtained with our method on meshes and point clouds from the proposed datasets. Notice how regions close to the boundary are still accurately matched despite the noisy descriptors.}
\end{figure*}

\noindent\textbf{Rank.}
In the first experiment we study the change in accuracy as the rank of the functional map is increased; this corresponds to using an increasing number of basis functions for the two shapes being matched.
For this experiment we compare with the baseline method of Ovsjanikov~\etal \cite{ovsjanikov12} by using the same dense descriptors as ours. 
 For fair comparisons, we did not impose map orthogonality or operator commutativity constraints~\cite{ovsjanikov12}, which cannot obviously be satisfied due to partiality. The results of this experiment are reported in Fig.~\ref{fig:rank}. As we can see from the plots, our method allows to obtain more accurate solutions as the rank increases, while an opposite behavior is observed for the other method.

\noindent\textbf{Representation.}
Our method is general enough to be applied to different shape representations, as long as a proper discretization of the Laplace operator is available. In Fig.~\ref{fig:examples} we show some qualitative examples of correspondences produced by our algorithm on simulated point clouds and depth maps. Here we use the method described in~\cite{belkin09} to construct a discrete Laplacian on the point clouds. This is traditionally considered a particularly challenging problem in robotics and vision applications, with few methods currently capable of giving satisfactory solutions without exploiting controlled conditions or domain-specific information (\eg, the knowledge that the shape being matched is that of a human). These are, to the best of our knowledge, the best results to be published so far for this class of problems.


\subsection{Comparisons}
We compared our method on the {\em cuts} and {\em holes} datasets (200 shape pairs in total)
; the results are shown in Fig.~\ref{fig:comparisons}.
As an additional experiment, we ran comparisons against \cite{ovsjanikov12} across increasing amounts of partiality. The rationale behind this experiment is to show that, at little or no partiality, our approach converges to the one described in \cite{ovsjanikov12}, currently among the state of the art in non-rigid shape matching. However, as partiality increases so does the sensitivity of the latter method. 
 Fig.~\ref{fig:horse_partiality} shows the results of this experiment.

 
Parameters for our method were chosen on the basis of the sensitivity analysis. Specifically, we used $k=100$ eigenfunctions per shape, and set $\mu_1 = \mu_3 = 1$, $\mu_4 = \mu_5 = 10^3$, and $\mu_2 = 10^2$. The different orders of magnitude for the $\mu$ coefficients are due to the fact that the regularizing terms operate at different scales. We also experimented with other values, but in all our tests we did not observe significant changes in accuracy.
Additional examples of partial matchings obtained with our method are shown in Fig.~\ref{fig:examples}.


\section{Discussion and conclusions}
\label{sec:conclusion}

In this paper we tackled the problem of dense matching of deformable shapes under partiality transformations. We cast our formulation within the framework of functional maps, which we adapted and extended to deal with this more challenging scenario. 
%
 Our approach is fully automatic and makes exclusive use of dense local features as a measure of similarity. Coupled with a robust prior on the functional correspondence derived from a perturbation analysis of the shape Laplacians, this allowed us to devise an effective optimization process with remarkable results on very challenging cases.
In addition to our framework for partial functional correspondence we also introduced two new datasets comprising hundreds of shapes, which we hope will foster further research on this challenging problem.

One of the main issues of our method concerns the existence of multiple optima, which is in turn related to the presence of non-trivial self-isometries on the considered manifolds. Since most natural shapes are endowed with intrinsic symmetries, one may leverage this knowledge in order to avoid inconsistent matchings. For example, including a smoothness prior on the correspondence might alleviate such imperfections and thus provide better-behaved solutions. 
Secondly, since the main focus of this paper is on tackling partiality rather than general deformations, our current formulation does not explicitly address the cases of topological changes and inter-class similarity (\eg, matching a man to a gorilla). However, the method can be easily extended to employ more robust descriptors such as pairwise features~\cite{rodola13,kaick13}, or to simultaneously optimize over ad-hoc functional bases on top of the correspondence. Finally, extending our approach to tackle entire shape collections, as opposed to individual pairs of shapes, represents a further exciting direction of research.

\section*{Acknowledgments}
The authors thank Matthias Vestner, Vladlen Koltun, Aneta Stevanovi\'{c}, Zorah L\"{a}hner, Maks Ovsjanikov, and Ron Kimmel for useful discussions. 
ER is supported by an Alexander von Humboldt Fellowship. 
MB is partially supported by the ERC Starting Grant No. 307047 (COMET).

\bibliographystyle{eg-alpha-doi}

\bibliography{egbib}

\small

\section*{Appendix A - Discretization}

We describe more in depth the discretization steps and give the implementation details of our algorithm (we skip trivial derivations for the sake of compactness). Detailed gradients for each term are given in Appendix B.

\paragraph*{Mesh parametrization.} 
We denote by $S$ a triangle mesh of $n$ points, composed of triangles $S_j$ for $j=1,\dots,m$. In the following derivations, we will consider the classical triangle-based parametrization described by the charts $x_j : \mathbb{R}^2 \to \mathbb{R}^3$
\begin{equation}
x_j(\alpha, \beta) = x_{j,1} + \alpha(x_{j,2}-x_{j,1}) + \beta(x_{j,3} - x_{j,1})\,,
\end{equation}
with $\alpha \in [0,1]$ and $\beta \in [0, 1-\alpha]$. With $x_{j,k} \in \mathbb{R}^3$ we denote the 3D coordinates of vertex $k \in \{1,2,3\}$ in triangle $S_j$.

Each triangle $S_j$ is equipped with a discrete metric tensor with coefficients
\begin{eqnarray}
g_j = \left( \begin{array}{cc} E_j&F_j \\ F_j&G_j \end{array} \right)\,,
\end{eqnarray}
where $E_j = \| x_{j,2}-x_{j,1} \|^2$, $F_j = \langle x_{j,2}-x_{j,1} , x_{j,3}-x_{j,1} \rangle$, and $G_j = \| x_{j,3}-x_{j,1} \|^2$. The volume element for the $j$-th triangle is then given by $\sqrt{\det g_j} = \sqrt{E_jG_j-F_j^2}$.

\paragraph*{Integral of a scalar function.}
Scalar functions $f:S \to \mathbb{R}$ are assumed to behave linearly within each triangle. Hence, $f(x(\alpha,\beta))$ is a linear function of $(\alpha,\beta)$ and it is uniquely determined by its values at the vertices of the triangle. The integral of $f$ over $S_j$ is then simply given by:
\begin{align}
&\int_0^1 \int_0^{1-\alpha} f(\alpha,\beta) \sqrt{\det g_j} d\beta d\alpha\\
&= \int\int f(0,0) (1-\alpha-\beta) + f(1,0) \alpha + f(0,1)\beta ~\sqrt{\det g_j} d\beta d \alpha\nonumber\\
&= \frac{1}{6}  (f(0,0) + f(1,0) + f(0,1))\sqrt{E_jG_j-F_j^2}\nonumber\\
&=  \frac{1}{3}  (f(0,0) + f(1,0) + f(0,1))\mathrm{area}(S_j) \label{eq:integral}\,,
\end{align}
where $f(0,0) = f(x_{j,1})$, $f(1,0) = f(x_{j,2})$, and $f(0,1) = f(x_{j,3})$.

\paragraph*{Gradient of a scalar function.}
For the intrinsic gradient of $f$ we get the classical expression in local coordinates:
\begin{eqnarray}
\nabla f =\left( \begin{array}{cc} \frac{\partial x_j}{\partial \alpha} & \frac{\partial x_j}{\partial \beta} \end{array} \right) \left( \begin{array}{cc} E_j&F_j \\ F_j&G_j \end{array} \right)^{-1} \left( \begin{array}{c} f_\alpha \\ f_\beta \end{array} \right)\,,
\end{eqnarray}
where we write $f_\alpha$ to denote the partial derivative $\frac{\partial f}{\partial \alpha} = f_{j,2} - f_{j,1}$ and similarly for $f_\beta$.  The norm of the intrinsic gradient over triangle $S_j$ is then given by:
\begin{align}
&\|\nabla f\|\\
&= \sqrt{\langle\nabla f, \nabla f\rangle}\nonumber\\
&= \sqrt{ \left( \begin{array}{cc} f_\alpha & f_\beta \end{array} \right) \left( \begin{array}{cc} E_j&F_j \\ F_j&G_j \end{array} \right)^{-1} \left( \begin{array}{c} f_\alpha \\ f_\beta \end{array} \right) }\nonumber\\
&= \sqrt{ \left( \begin{array}{cc} f_\alpha & f_\beta \end{array} \right) \left( \begin{array}{cc} G_j&-F_j \\ -F_j&E_j \end{array} \right) \left( \begin{array}{c} f_\alpha \\ f_\beta \end{array} \right) } \frac{1}{\sqrt{\det g_j}}\nonumber\\
&= \sqrt{\frac{f_\alpha^2 G_j - 2 f_\alpha f_\beta F_j + f_\beta^2 E_j}{\det g_j} }\,.
\end{align}
Note that, since we take $f$ to be linear, the gradient $\nabla f$ is constant within each triangle. 
We can then integrate $\nabla f$ over $S_j$  as follows:
\begin{align}
&\int_{S_j} \| \nabla f(x)\| \sqrt{\det g_j}  d\alpha d\beta\\
&= \int_{S_j} \sqrt{\frac{f_\alpha^2 G_j - 2 f_\alpha f_\beta F_j + f_\beta^2 E_j}{\det g_j} } \sqrt{\det g_j}d\alpha d\beta\nonumber\\
&= \int_{S_j} \sqrt{f_\alpha^2 G_j - 2 f_\alpha f_\beta F_j + f_\beta^2 E_j } ~d\alpha d\beta\nonumber\\
&= \frac{1}{2}  \sqrt{f_\alpha^2 G_j - 2 f_\alpha f_\beta F_j + f_\beta^2 E_j }\label{eq:gradient}\,.
\end{align}

In the following, we write $\mathcal{N}$ and $\mathcal{M}$ to denote the partial and full shape respectively. Further, let $\{ \lambda_i^\mathcal{N} \}_{i=1,\dots,k}$ be the first $k$ eigenvalues of the Laplacian on $\mathcal{N}$, and similarly for $\{ \lambda_i^\mathcal{M} \}_{i=1,\dots,k}$. The functional map $\vct{C}$ has size $k \times k$.
%
%
%
\paragraph*{Mumford-Shah functional ($\mu_2$-term).}
Following Equations \eqref{eq:integral} and \eqref{eq:gradient}, we immediately obtain:
\begin{align}
&\int_S \xi(v) \| \nabla v \| dx\nonumber\\
&= \sum_{j=1}^m \int_{S_j} \xi(\alpha,\beta) \| \nabla v \| \sqrt{\det g_j} d\beta d\alpha\nonumber\\
&= \sum_{j=1}^m \sqrt{v_\alpha^2 G_j - 2 v_\alpha v_\beta F_j + v_\beta^2 E_j } \int_{S_j} \xi(\alpha,\beta) d\beta d\alpha\nonumber\\
&\approx \frac{1}{6} \sum_{j=1}^m \sqrt{v_\alpha^2 G_j - 2 v_\alpha v_\beta F_j + v_\beta^2 E_j }  (\xi(0,0) + \xi(1,0) + \xi(0,1))\nonumber\,,
\end{align}
where $\xi(0,0) = \xi(v(x_{j,1}))$, $\xi(1,0) = \xi(v(x_{j,2}))$, and $\xi(0,1) = \xi(v(x_{j,3}))$.

\paragraph*{Weight matrix ($\mu_3$-term).}
Recall from Section \ref{sec:perturb} and Figure \ref{fig:spectra} that an estimate for the rank of $\vct{C}$ can be easily computed as
\begin{equation}\label{eq:rank}
r = \max \{ i ~|~ \lambda_i^\mathcal{N} < \max_j \lambda_j^\mathcal{M} \}\,.
\end{equation}
We use this information in order to construct the weight matrix $\vct{W}$, whose diagonal slope directly depends on $r$.

To this end, we model $\vct{W}$ as a regular $k \times k$ grid in $\mathbb{R}^2$. The slanted diagonal of $\vct{W}$ is a line segment $\vct{\delta}(t) = \vct{p} + t  \frac{\vct{n}}{\| \vct{n} \|}$ with $t \in \mathbb{R}$, where $\vct{p}=(1,1)\T$ is the matrix origin, and $\vct{n} = (1, r/k)\T$ is the line direction with slope $r/k$.
The high-frequency spread in $\vct{C}$ is further accounted for by funnel-shaping $\vct{W}$ along the slanted diagonal. We arrive at the following expression for $\vct{W}$:
\begin{equation}\label{eq:wij}
w_{ij} = e^{-\sigma\sqrt{i^2 + j^2}} \| \frac{\vct{n}}{\| \vct{n} \|} \times ((i,j)\T - \vct{p}) \| \,,
\end{equation}
where the second factor is the distance from the slanted diagonal $\vct{\delta}$, and $\sigma \in \mathbb{R}_+$ regulates the spread around $\vct{\delta}$. In our experiments we set $\sigma = 0.03$.


\paragraph*{Orthogonality ($\mu_4 , \mu_5$-terms).}
For practical reasons, we incorporate the off-diagonal and diagonal terms within one term with the single coefficient $\mu_{4,5}$. In addition, we rewrite the off-diagonal penalty using the following equivalent expression:
\begin{equation}
\sum_{i \neq j} ( \mathbf{C}^\mathrm{T} \mathbf{C} )^2_{ij} = \| \vct{C}\T\vct{C} \|_\mathrm{F}^2 - \sum_i (\vct{C}\T\vct{C})^2_{ii}\,.
\end{equation}
Vector $\vct{d} \in \mathbb{R}^k$ is constructed by setting the first $r$ elements (according to \eqref{eq:rank}) equal to 1, and the remaining $k-r$ elements equal to 0.

\section*{Appendix B - Gradients}

We find local solutions to each optimization problem by the (nonlinear) conjugate gradient method. In this Section we give the detailed gradient derivations of all terms involved in the optimization. 

In order to keep the derivations practical, we will model function $v$ by its corresponding $n$-dimensional vector $\vct{v}$.
Note that, depending on the optimization step, the gradients are computed with respect to either $\vct{v}$ or $\vct{C}$.

\paragraph*{Data term (w.r.t. ${v}$).}
Let $q$ denote the number of corresponding functions between the two shapes. Matrices $\F$ and $\G$  contain the column-stacked functions defined over $\N$ and $\M$ and have size $n\times q$. The respective projections onto the corresponding functional spaces $\bm{\Phi}$ and $\bm{\Psi}$ are stored in the $k\times q$ matrices $\A$ and $\B$ respectively. 
Let us write $\D_{ij}$ to identify the elements of the matrix $\vct{CA} - \vct{B} (\eta(v))$, we then have
\begin{align}
&\frac{\partial}{\partial v_p} \| \vct{\C\A} - \vct{\B}(\eta(v))\|_{2,1}\nonumber\\
&= \frac{\partial}{\partial v_p} \sum_{j=1}^q \left( \sum_{i=1}^n \D_{ij}^2 \right)^\frac{1}{2} \nonumber\\
&= \sum_{j=1}^q \left( \sum_{i=1}^n \D_{ij}^2 \right)^{-\frac{1}{2}} \sum_{i=1}^n \D_{ij} \frac{\partial}{\partial v_p} \D_{ij}\,. \label{eq:derdatatermvp}
\end{align}
Since $\B (\eta(v))_{ij} = \sum_k^n \bm{\Psi}^\mathrm{T}_{ik} \eta(v_k) \F_{kj}$, we have:
\begin{equation}
\frac{\partial}{\partial v_p} \D_{ij} = \frac{\partial}{\partial v_p} [\C\A_{ij} - \B_{ij}]
= \bm{\Psi}^\mathrm{T}_{ip} \F_{pj}  \frac{\partial}{\partial v_p}\eta(v_p)\,.
\end{equation}
Finally:
\begin{equation}\label{eq:gradetav}
\frac{\partial}{\partial v_p}\eta(v_p) = 1-\tanh^2(2 v_p -1).
\end{equation}

\paragraph*{Area term ($\mu_1$-term w.r.t. ${v}$).}
The derivative of the discretized area term is:
\begin{align*}
&\frac{\partial}{\partial v_p} \left( \sum_{i=1}^n (S_\N)_i - \sum_{i=1}^n (S_\M)_i \eta(v_i) \right)^2\\
&= 2 \left( \sum_{i=1}^n (S_\N)_i - \sum_{i=1}^n (S_\M)_i \eta(v_i) \right) S_M \frac{\partial}{\partial v_p}\eta(v_p)
\end{align*}
where $(S_\M)_i$ and $(S_\N)_i$ are the local area elements associated with the $i$-th vertex of meshes $\M$ and $\N$ respectively. For the derivative of $\eta(v_p)$ see equation \eqref{eq:gradetav}.

\paragraph*{Mumford-Shah functional ($\mu_2$-term w.r.t. ${v}$).}
Computing the gradient $\nabla_\vct{v} \int_S \xi(\vct{v}) \| \nabla \vct{v} \| dx$ involves computing partial derivatives of $\xi(\vct{v})$ with respect to $\vct{v}$. These are simply given by:
\begin{align*}
\frac{\partial}{\partial v_k} \xi (v_k) &= \frac{\partial}{\partial v_k} e^{-\frac{\tanh(2 v_k - 1)}{4 \sigma^2}}\\
&= -\frac{1 - \tanh^2(2 v_k -1)}{2 \sigma^2} e^{-\frac{\tanh(2 v_k - 1)}{4 \sigma^2}}\,.
\end{align*}
In the following derivations we set $D_j \equiv \sqrt{v_\alpha^2 G_j - 2 v_\alpha v_\beta F_j + v_\beta^2 E_j }$, and $D_j =0$ whenever $\nabla \vct{v} = \vct{0}$.
The gradient of the Mumford-Shah functional is then composed of the partial derivatives:
\begin{align*}
&\frac{\partial}{\partial v_k}  \int_S \xi(\vct{v}) \| \nabla \vct{v} \| dx\nonumber\\
&= \sum_{j=1}^m \frac{\partial}{\partial v_k} \int_{S_j} \xi(\vct{v}) \| \nabla \vct{v} \| \nonumber\\
&= \frac{1}{6} \sum_{j \in N(k)} \frac{\partial}{\partial v_k}  D_j ( \xi(v_k) + \xi( v_{j,2} ) + \xi ( v_{j,3} ) )\nonumber\\
&= \frac{1}{6} \sum_{j \in N(k)}  ( \xi(v_k) + \xi( v_{j,2} ) + \xi ( v_{j,3} ) ) \frac{\partial}{\partial v_k}D_j + D_j \frac{\partial}{\partial v_k} \xi(v_k)\nonumber\\
&= \frac{1}{6} \sum_{j \in N(k)}  ( \xi(v_k) + \xi( v_{j,2} ) + \xi ( v_{j,3} ) ) \frac{1}{2D_j}  \frac{\partial}{\partial v_k} D_j^2 + D_j  \frac{\partial}{\partial v_k} \xi(v_k) \nonumber\\
&=\frac{1}{6} \sum_{j \in N(k)}  ( \xi(v_k) + \xi( v_{j,2} ) + \xi ( v_{j,3} ) ) K_j  + D_j \frac{\partial}{\partial v_k} \xi(v_k) \nonumber
\end{align*}
where we write $K_j \equiv \frac{1}{D_j} ((v_k - v_{j,2})(G_j-F_j)+(v_k-v_{j,3})(E_j-F_j))$, and  $j \in N(k)$ are the indices of the triangles containing the $k$-th vertex. Note that we slightly abuse notation by writing $v_k$, $v_{j,2}$, and $v_{j,3}$ to denote the three vertices of the $j$-th triangle, even though in general the ordering might be different depending on the triangle.

\paragraph*{Data term (w.r.t. C).}
The derivative of the data term with respect to $\C$ is similar to \eqref{eq:derdatatermvp}. The only difference is in the partial derivative:
\begin{equation}
\frac{\partial}{\partial \C_{pq}} \D_{ij} = \frac{\partial}{\partial \C_{pq}} \sum_{i=1}^k \C_{ik} \A_{kj} = 
\begin{cases}  \C_{pq} \A_{qj} & \mbox{if } i = p  \\
                             0 & \mbox{otherwise.}\end{cases} 
\end{equation}

\paragraph*{Weight matrix ($\mu_3$-term w.r.t. C).}
This is simply given by:
\begin{equation}
\frac{\partial}{\partial \C_{pq}} \| \C \circ \vct{W} \|^2_\mathrm{F} = \C_{pq} (\vct{W}_{pq})^2\,.
\end{equation}

\paragraph*{Orthogonality ($\mu_{4,5}$-term w.r.t. C).}
The gradient of the last term can be finally obtained as:
\begin{equation*}
\begin{aligned}
&\frac{\partial}{\partial \C_{pq}} 
\left[ \| \C\T\C \|_\mathrm{F}^2 - \sum_i (\vct{C}\T\vct{C})^2_{ii}\
 + \sum_i ((\vct{C}\T\vct{C})_{ii} - d_i)^2 \right]
\\& \; = 4(\C \C^T \C)_{pq} + 2 \sum_{i} \left[ -\sum_k \C_{ki}^2 \frac{\partial}{\partial \C_{pq}} \sum_k \C_{ki}^2 +(\sum_k \C_{ki}^2 - d_i)\frac{\partial}{\partial \C_{pq}} \sum_k \C_{ki}^2 \right]
\\& \; = 4[ (\C \C^T \C)_{pq} - d_q \C_{pq}]\,.
\end{aligned}
\end{equation*}

\section*{Appendix C - Perturbation Analysis}

\newtheorem{thm}{Theorem}
\begin{thm}
Let $\vct{L}_\mathcal{N}+t\vct{P}_\mathcal{N} = \bm{\Phi}(t)\T \bm{\Lambda}(t) \bm{\Phi}(t)$, where $\bm{\Lambda}(t) = \mathrm{diag}(\lambda_1(t), \hdots, \lambda_n(t))$ is a diagonal matrix of eigenvalues, and $\bm{\Phi}(t)$ are the corresponding eigenvectors. 
The derivative of the non-trivial eigenvalues is given by 
\begin{equation}\label{eq:eigd}
 \frac{d}{dt}\lambda_i = \sum_{v,w\in \partial \mathcal{N}} (\vct{P}_\mathcal{N})_{vw} {\phi}_{iv} {\phi}_{iw}
 = \bm{\phi}_i^\top \vct{P}_\mathcal{N} \bm{\phi}_i. 
\end{equation}
\end{thm}
{\em Proof:} Let $\vct{A}(t)$ be a symmetric real $n\times n$ matrix  parametrized by $t \in T\subseteq\mathbb{R}$,  with $\bm{\Phi}(t)$ and $\bm{\Lambda}(t)$ being the eigenvector and eigenvalue matrices, {\em i.e.}, for all $t \in T$ we have 
\begin{equation}\label{eq:eigen}
\vct{A}(t) \bm{\Phi}(t) = \bm{\Phi}(t)\bm{\Lambda}(t)
\end{equation}
and $\bm{\Lambda}(t)$ is diagonal and $\bm{\Phi}(t)$ orthogonal.

Following~\cite{NME:NME1620260202}, if all the eigenvalues are distinct, then we can compute the derivatives of the eigenvalues at $t=0$ as
\begin{equation}
 \lambda^\prime_i = \bm{\phi}_i\T \vct{A}^\prime \bm{\phi}_i
\end{equation}
where $\vct{A}^\prime$, the derivative of $\vct{A}(t)$, and the eigenvectors $\bm{\phi}_i$ are considered at $t=0$.
In fact, differentiating (\ref{eq:eigd}), we obtain
\begin{equation}\label{eq:diff}
\vct{A}^\prime \bm{\Phi} + \vct{A} \bm{\Phi}^\prime = \bm{\Phi}^\prime \bm{\Lambda} + \bm{\Phi} \bm{\Lambda}^\prime\,.
\end{equation}
Left-multiplying both sides by $\bm{\Phi}\T$, setting $\bm{\Phi}^\prime=\bm{\Phi}\vct{B}$ for a matrix $\vct{B}$ to be determined, and recalling that $\bm{\Phi}\T \vct{A} \bm{\Phi}=\bm{\Lambda}$, we have
\begin{equation}\label{eq:diff2}
\bm{\Phi}\T \vct{A}^\prime \bm{\Phi} + \bm{\Lambda}\vct{B} = \vct{B}\bm{\Lambda} + \bm{\Lambda}^\prime\,,
\end{equation}
from which
\begin{equation}
\diag(\bm{\Lambda}^\prime) = \diag(\bm{\Phi}\T \bm{A}^\prime \bm{\Phi}) + \diag(\bm{\Lambda}\vct{B} - \vct{B}\bm{\Lambda}) = \diag(\bm{\Phi}\T \vct{A}^\prime \bm{\Phi})\,.
\end{equation}

Going back to our case, we take the simplifying assumptions that $\vct{L}_\mathcal{M}$ and $\vct{L}_\mathcal{N}$ do not have repeated eigenvalues. Let
\begin{eqnarray}
 \vct{L}_\mathcal{N} &=& \bm{\Phi}\T \bm{\Lambda} \bm{\Phi}\\
 \vct{L}_{\overline{\mathcal{N}}} &=& \overline{\bm{\Phi}}\T \overline{\bm{\Lambda}} \overline{\bm{\Phi}}
\end{eqnarray}
be the spectral decompositions of $\vct{L}_\mathcal{N}$ and $\vct{L}_{\overline{\mathcal{N}}}$ respectively. According to the previous result, we can write the derivative of $\lambda_i$ eigenvalue of $\vct{L}_\mathcal{N}$ and, thus, of $\vct{L}(0)$, as:
\begin{equation}
 \lambda_i^\prime = \bm{\phi}_i\T \vct{P}_\mathcal{N} \bm{\phi}_i = \sum_{v,w\in \partial \mathcal{N}} (\vct{P}_\mathcal{N})_{vw} {\phi}_{iv} {\phi}_{iw} \,.
\end{equation}

\begin{thm}
Assume that $\vct{L}_{{\mathcal{N}}}$ has distinct eigenvalues ($\lambda_i \neq \lambda_j$ for $i\neq j$), and furthermore, the non-zero eigenvalues are all distinct from the eigenvalues of $\vct{L}_{\overline{\mathcal{N}}}$ ($\lambda_i \neq \overline{\lambda}_j$ for all $i, j$). 
Let $\vct{L}_\mathcal{N}+t\vct{P}_\mathcal{N} = \bm{\Phi}(t)\T \bm{\Lambda}(t) \bm{\Phi}(t)$, where $\bm{\Lambda}(t) = \mathrm{diag}(\lambda_1(t), \hdots, \lambda_n(t))$ is a diagonal matrix of eigenvalues, and $\bm{\Phi}(t)$ are the corresponding eigenvectors.
Then, the derivative of the non-constant eigenvector 
is given by 
\begin{equation}
 \frac{d}{dt}\bm{\phi}_i = \sum_{ {j=1}\atop{j\neq i}}^{n}  \frac{\bm{\phi}_i\T \vct{P}_\mathcal{N} \bm{\phi}_j}{{\lambda}_i-{\lambda}_j} \bm{\phi}_j
 + \sum_{j=1}^{\overline{n}}  \frac{\bm{\phi}_i\T \vct{P}\; \overline{\bm{\phi}}_j}{{\lambda}_i-\overline{{\lambda}}_j} \overline{\bm{\phi}}_j\,.
\end{equation}
\end{thm}
{\em Proof:}
Under the same assumptions as for the previous theorem, from (\ref{eq:diff2}) we have
\begin{eqnarray}
(\bm{\Phi}\T \vct{A}^\prime \bm{\Phi})_{ij} + (\bm{\Lambda}\vct{B})_{ij} &=& (\vct{B}\bm{\Lambda})_{ij} + (\bm{\Lambda}^\prime)_{ij}\\
\bm{\phi}_i\T \vct{A}^\prime \bm{\phi}_j + \lambda_i b_{ij} &=& b_{ij}\lambda_j + 0\,,
\end{eqnarray}
from which
\begin{equation}\label{eq:mix}
 b_{ij} = \frac{\bm{\phi}_i\T \vct{A}^\prime \bm{\phi}_j}{\lambda_j-\lambda_i}\,.
\end{equation}
due to the orthogonality of $\bm{\Phi}(t)$ we have that $\vct{B}$ is skew-symmetric, and thus, $b_{ii}=0$.
From the relation $\bm{\Phi}^\prime = \bm{\Phi}\vct{B}$ we obtain
\begin{equation}
  \frac{d}{dt}\bm{\phi}_i =  \sum_{j\neq i} b_{ji} \bm{\phi}_j = \sum_{j\neq i} \frac{\bm{\phi}_j\T A^\prime \bm{\phi}_i}{\lambda_i-\lambda_j} \bm{\phi}_j\,.
\end{equation}

Going back to our case, recall that the set of eigenvalues of $\vct{L}(0)$ is the union of the eigenvalues of $\vct{L}_\mathcal{N}$ and $\vct{L}_{\overline{\mathcal{N}}}$ and the corresponding eigenvectors are obtained from those of $\vct{L}_\mathcal{N}$ and $\vct{L}_{\overline{\mathcal{N}}}$ by padding with zeros on the missing parts.
Denoting with $\lambda_i$ and $\bm{\phi}_i$ the eigenvalues and corresponding eigenvectors of $\vct{L}_\mathcal{N}$, and $\overline{{\lambda}}_j$ and $\overline{\bm{\phi}}_j$ the eigenvalues and corresponding eigenvectors of $\vct{L}_{\overline{\mathcal{N}}}$, we have
\begin{equation}
 \frac{d}{dt}\bm{\phi}_i = \sum_{ {j=1}\atop{j\neq i}}^{n}  \frac{\bm{\phi}_i\T \vct{P}_\mathcal{N} \bm{\phi}_j}{{\lambda}_i-{\lambda}_j} \bm{\phi}_j
 + \sum_{j=1}^{\overline{n}}  \frac{\bm{\phi}_i\T \vct{P}\; \overline{\bm{\phi}}_j}{{\lambda}_i-\overline{{\lambda}}_j} \overline{\bm{\phi}}_j\,.
\end{equation}

\paragraph*{Boundary interaction strength.}
We can measure the variation of the eigenbasis as a function of the boundary $\mathcal{B}$ splitting $\mathcal{M}$ into $\mathcal{N}$ and $\overline{\mathcal{N}}$ as
\begin{eqnarray}
 \partial \bm{\Phi}(\mathcal{B}) &=& \sum_{i=1}^{n} \| \bm{\phi}_i^\prime \|^2_{\mathcal{N}} 
= \sum_{i=1}^{n}  \left( \sum_{j=1\atop{j\neq i}}^{n} \frac{\bm{\phi}_i\T \vct{D}_\mathcal{N} \bm{\phi}_j}{{\lambda}_i-{\lambda}_j} \right)^2.
\end{eqnarray}
Let us now consider the function:
\begin{equation}
f(v) =  \sum_{{i,j=1}\atop{j\neq i}}^{n} \left(\frac{{\phi}_{iv}{\phi}_{jv}}{{\lambda}_i-{\lambda}_j}\right)^2\,.
\end{equation}
Assuming $\vct{D}_\mathcal{N}$ diagonal and with constant diagonal elements $k$, we have 
\begin{equation}
 k \int_{\mathcal{B}} f(v)\,dv \geq  \partial \bm{\Phi}(\mathcal{B})\,,
\end{equation}
in fact:
\begin{eqnarray}
\partial \bm{\Phi}(\mathcal{B}) 
&\approx& k \sum_{i=1}^{n}  \left( \sum_{j=1\atop{j\neq i}}^{n}\frac{\sum_{v\in \partial \mathcal{M}}{\phi}_{iv}\ {\phi}_{jv}}{{\lambda}_i-{\lambda}_j} \right)^2 \nonumber\\
&\leq& k \sum_{v\in \partial \mathcal{M}}  \sum_{{i,j=1}\atop{j\neq i}}^{n} \left( \frac{{\phi}_{iv}\ {\phi}_{jv}}{{\lambda}_i-{\lambda}_j} \right)^2 = k \sum_{v\in \partial \mathcal{M}} f(v).\,
\end{eqnarray}

\end{document}